\documentclass{article}

 \usepackage[main, final, nonatbib]{neurips_2026}

\usepackage[utf8]{inputenc} 
\usepackage[T1]{fontenc}    
\usepackage{url}            
\usepackage{booktabs}       
\usepackage{amsfonts}       
\usepackage{nicefrac}       
\usepackage{microtype}      
\usepackage[table]{xcolor}  

\usepackage[noadjust]{cite}

\renewcommand{\citepunct}{],\penalty\citepunctpenalty\ [}
\renewcommand{\citedash}{]--[}

\makeatletter
\def\@compress@cite#1#2#3{%
  \ifnum\@cite@incr=\z@
    \advance\@tempcnta\@cclvi
    \ifnum #1=\@tempcnta
       \def\@h@ld{\citedash\@cite@out{#3}}%
       \mathchardef\@cite@incr=\@cclvi
    \else
       \advance\@tempcnta-\@cclv
       \ifnum #1=\@tempcnta
         \def\@h@ld{\citedash\@cite@out{#3}}%
         \mathchardef\@cite@incr=\@ne
       \else
         \@citea \@cite@out{#3}%
       \fi
    \fi
  \else
    \advance\@tempcnta\@cite@incr
    \ifnum #1=\@tempcnta
       \def\@h@ld{\citedash \@cite@out{#3}}%
    \else
       \@h@ld \@citea \@cite@out{#3}%
       \let\@h@ld\@empty
       \mathchardef\@cite@incr=\z@
    \fi
  \fi
  \@tempcnta#1\let\@citea\citepunct
}
\makeatother

\providecommand{\citep}{\cite}
\providecommand{\citet}{\cite}

\title{\kpwm: Control-Oriented Structured Bilinear World Models under Partial Observation}

\author{%
  Santosh M. Rajkumar\thanks{ \tiny Department of Mechanical \& Aerospace Engineering (in Columbus, OH, USA)  \big (E-mail: \texttt{\{rajkumar.36,\ goswami.78\}@osu.edu} \big )} \\
  The Ohio State University
  \And
  Sriram Narayanan\thanks{ \tiny Department of Mechanical, Materials, and Aerospace Engineering (in Morgantown, WV, USA) (E-mail: \texttt{sriram.narayanan@mail.wvu.edu})}\\
  West Virginia University\\
  \And
  Samuel E. Otto\thanks{\tiny Sibley School of Mechanical and Aerospace Engineering (in Ithaca, NY, USA) (E-mail: \texttt{s.otto@cornell.edu}) \\
  \hspace{10em}  \href{https://github.com/santoshrajkumar/k-pwm-world-model-control}{%
  \lightbutton{\faGithub}{ \tiny \textbf{Open-source Code}}%
}
  }\\
  Cornell University\\
  \And
  Debdipta Goswami\footnotemark[1]\\
  The Ohio State University
}

\usepackage{amsmath,amssymb,mathtools}
\usepackage{dsfont}
\usepackage[ruled,vlined,linesnumbered]{algorithm2e}
\usepackage{enumitem}
\usepackage{graphicx}
\usepackage{multirow}
\usepackage{pifont}
\usepackage{subcaption}
\usepackage{mathrsfs}
\usepackage{amsthm}
\usepackage[
    colorlinks=true,
    linkcolor=blue,
    citecolor=magenta,
    urlcolor=blue
]{hyperref}
\usepackage[nameinlink]{cleveref}

\usepackage[table]{xcolor}
\definecolor{heatamber}{HTML}{F2B134}
\newcommand{\heatcell}[2]{%
  \cellcolor{heatamber!#1}$\scriptstyle #2$}
\usepackage{caption} %

\usepackage[most]{tcolorbox}
\usepackage{fontawesome5}

\newcommand{\lightbutton}[2]{%
  \tcbox[
    on line,
    colback=gray!8,
    colframe=gray!25,
    boxrule=0.4pt,
    arc=2pt,
    boxsep=0pt,
    left=6pt, right=6pt,
    top=2pt, bottom=2pt
  ]{\small #1\hspace{0.4em}#2}%
}
  
\crefname{section}{\S}{\S\S}
\Crefname{section}{\S}{\S\S}
\crefname{subsection}{\S}{\S\S}
\Crefname{subsection}{\S}{\S\S}
\crefname{subsubsection}{\S}{\S\S}
\Crefname{subsubsection}{\S}{\S\S}
\crefname{figure}{Fig.}{Figs.}
\Crefname{figure}{Fig.}{Figs.}
\crefname{table}{Table}{Tables}
\Crefname{table}{Table}{Tables}
\crefname{equation}{Eq.}{Eqs.}
\Crefname{equation}{Eq.}{Eqs.}
\crefname{algorithm}{Alg.}{Algs.}
\Crefname{algorithm}{Alg.}{Algs.}
\crefname{algocf}{Alg.}{Algs.}
\Crefname{algocf}{Alg.}{Algs.}
\crefname{assumption}{Assumption}{Assumptions}
\Crefname{assumption}{Assumption}{Assumptions}
\crefname{theorem}{Theorem}{Theorems}
\Crefname{theorem}{Theorem}{Theorems}
\crefname{corollary}{Corollary}{Corollaries}
\Crefname{corollary}{Corollary}{Corollaries}
\crefname{proposition}{Proposition}{Propositions}
\Crefname{proposition}{Proposition}{Propositions}
\crefname{lemma}{Lemma}{Lemmas}
\Crefname{lemma}{Lemma}{Lemmas}
\crefname{definition}{Definition}{Definitions}
\Crefname{definition}{Definition}{Definitions}
\crefname{remark}{Remark}{Remarks}
\Crefname{remark}{Remark}{Remarks}

\newtheorem{assumption}{Assumption}

\newtheorem{proposition}{Proposition}

\newtheorem{remark}{Remark}
\usepackage{wrapfig}

\newcommand{\kpwm}{\ensuremath{\mathrm{K\text{-}PWM}}}

\newcommand{\R}{\mathbb{R}}

\newcommand{\E}{\mathbb{E}}

\newcommand{\N}{\mathbb{N}}
\newcommand{\Nc}{\mathcal{N}}

\newcommand{\Kc}{\mathcal{K}}

\newcommand{\kron}{\otimes}

\DeclareMathOperator*{\trace}{tr}

\newcommand{\dx}{\mathrm{d}}

\def\bs{\boldsymbol}
\def\mf{\mathbf}

\def\mc{\mathcal}

\def\mrm{\mathrm}

\def\mfk{\mathfrak}

\def\what{\widehat}
\def\mcr{\mathscr}

\newcommand{\paren}[1]{\left( #1 \right)}
\newcommand{\brak}[1]{\left[ #1 \right]}
\newcommand{\set}[1]{\left\{ #1 \right\}}

\newcommand{\bmat}[1]{\begin{bmatrix}#1\end{bmatrix}}

\newcommand{\Aref}[1]{\Cref{#1}}

\newcommand{\Fref}[1]{\Cref{#1}}
\newcommand{\Pref}[1]{\Cref{#1}}

\newcommand{\Sref}[1]{\cref{#1}}

\newcommand{\alnn}[1]{\begin{align}#1\end{align}}

\newcommand{\salnn}[2]{%
\begin{subequations}\label{#1}
\begin{align}
#2
\end{align}
\end{subequations}
}

\usepackage{titletoc}

\begin{document}

\maketitle

\begin{abstract}

World models learn internal simulators of environment dynamics for prediction, planning, and control, yet learning them from limited, noisy, and partial observations remains challenging. While much progress has focused on pixel-based observations, many physical systems instead provide vector-valued measurements that reveal only part of the underlying state. We introduce \kpwm{}, a probabilistic world model for prediction and control under partial observation. \kpwm{} couples Koopman-bilinear latent dynamics with a nonlinear observation decoder and performs probabilistic latent-state inference through the learned model via extended Kalman filtering and smoothing, without a separately trained inference network. Its bilinear structure enables exact propagation of the maintained Gaussian belief through the latent transition and efficient rollouts for uncertainty-aware model predictive path integral (MPPI) control. For training \kpwm{}, we present a generalized expectation conditional maximization (GECM) procedure with closed-form updates for the generators, noise covariances, and latent prior, complemented by gradient-based decoder learning. A Koopman-consistent initialization selects the latent dimension from data and initializes the action-dependent dynamics. With limited training data, \kpwm{} achieves lower long-horizon prediction error than the evaluated baselines on partially observed Gymnasium MuJoCo systems, while substantially reducing training time relative to the strongest prediction baseline. On high-dimensional MuJoCo Menagerie robots, it achieves competitive tracking accuracy with low per-step computation times, demonstrating potential for real-time deployment. These results support structured probabilistic dynamics as a data- and compute-efficient approach to world modeling and control under partial observation.

\end{abstract}

\section{Introduction}\label{sec:intro}

Physical AI systems, such as robots and other embodied agents, must infer environmental dynamics from noisy, partial observations and anticipate the consequences of their actions \cite{kirchner2026survey}. Although physics-based models and system identification underpin prediction and control, modeling complex, contact-rich, and unstructured environments remains challenging, and large-scale models can impose substantial computational demands \cite{kirchner2026survey,jiang2022data,benner2015survey}. World models (WMs) learn compact, predictive representations of environmental dynamics, supporting forward prediction and counterfactual rollouts for decision-making \cite{lecun2022path,li2025comprehensive}. Unlike model-free RL (MFRL), which learns policies or value functions without explicitly modeling dynamics, WMs predict action outcomes to support model predictive control (MPC) and model-based RL (MBRL) \cite{sutton1991dyna,ding2025understanding}. 
Compared with MFRL, WMs can reduce costly interactions with physical systems by enabling planning and learning through imagined rollouts \cite{wu2023daydreamer}.

However, learning reliable WMs from noisy, partial observations remains challenging in real-world robotics \cite{Platt-RSS-10,ordonez2017learning,li2025uncertainty}. Partial observation affects both pixel- and vector-based settings \cite{huang2026vidworld,deihim2025transformer}, as sensor measurements often reveal only a subset of the underlying state \cite{asadi2022bayesian,ozan2025data,rajkumar2026adp}. For example, a mobile robot may observe its position without directly measuring its velocity. The resulting observation sequences can be non-Markovian even when the underlying dynamics are Markovian, motivating probabilistic latent-state inference to maintain a belief over hidden states and propagate uncertainty through time.

Existing WM learning methods often require substantial action-labeled data \cite{gao2025adaworld,huang2026vidworld}, posing challenges for learning directly from limited environmental interactions. A common approach is to learn models in simulation and subsequently adapt them to the target environment; however, data-efficient adaptation to changing environments remains challenging \cite{wu2023daydreamer}. These challenges motivate methods that can learn WMs directly from limited interaction data. Parametric state-space models (SSMs) that incorporate dynamical or physical structure offer a promising approach to improving data efficiency and generalization while supporting reliable continuous control \cite{lutter2018deep,gupta2020structured}. Moreover, interpretable latent dynamics can facilitate failure diagnosis and monitoring of robotic systems during deployment \cite{zhang2026factdmpc}.  Together, these considerations motivate investigating structured parametric SSMs for data-efficient and interpretable world modeling.

Real-world robotic tasks subject to actuator limits and environmental constraints often require rapid and continuous control updates, motivating the use of real-time MPC \cite{lee2023realtime,neunert2018wholebody,venkatesh2023sampling}. MPC is also used in MBRL to select actions by optimizing action sequences online in the latent space of learned WMs \cite{hansen2022temporal,hansen2024tdmpc}. However, the high-dimensional action spaces of complex robotic systems can make online planning computationally demanding, limiting scalability and real-time deployment \cite{bertsekas2019multiagent,sferrazza2024humanoidbench}. This limitation is further exacerbated in WMs that use densely connected neural networks \cite{zhang2026factdmpc}.  Introducing control-oriented dynamical structure into latent transitions can facilitate efficient predictive control, as illustrated by locally linear SSMs \cite{watter2015embed,banijamali2018robust}. This motivates the development of control-oriented WMs whose structured latent transitions enable efficient planning for real-time deployment.

In this work, we develop \kpwm{}, a \textbf{K}oopman bilinear-structured \textbf{P}robabilistic \textbf{W}orld \textbf{M}odel for prediction and downstream control of partially observed systems with vector-valued measurements. Our key contributions are:
i) \textbf{Probabilistic world modeling with structured latent dynamics and model-induced inference:} we couple Koopman-bilinear latent dynamics with a nonlinear decoder and model-induced inference, without a separately trained inference network. Koopman coordinates are treated as latent random variables inferred through the observation likelihood rather than lifted from full-state measurements, and for fixed actions the linear-Gaussian transition propagates the maintained Gaussian belief exactly, with linearization confined to the decoder.
ii) \textbf{Structure-exploiting GECM learning:} we derive closed-form conditional updates for the generators, noise covariances, and initial latent-state distribution, complemented by gradient-based nonlinear decoder learning.
iii) \textbf{Koopman-consistent initialization and dimension selection:} we combine delay-coordinate representations with action-conditioned generator regression to initialize the latent dynamics and select the latent dimension from data.
iv) \textbf{Uncertainty-aware planning and empirical validation:} we integrate \kpwm{} into an uncertainty-aware MPPI scheme and demonstrate improved prediction performance from limited data and reduced training time on partially observed Gymnasium MuJoCo systems, alongside competitive tracking accuracy and computationally efficient control on four high-dimensional MuJoCo Menagerie robots.

\subsection{Related Work}

\paragraph{WMs with pixel-based observations.} Early pixel-based WMs include \cite{ha2018recurrent}, where a VAE
compresses images into latent codes and an RNN predicts latent dynamics. PlaNet learns latent dynamics for planning \cite{hafner2019learning}, while the Dreamer family trains policies through latent imagination, spanning continuous
control \cite{hafner2019dreamer}, Atari with discrete latents
\cite{hafner2021mastering}, real-robot online learning \cite{wu2023daydreamer}, and scalable control across diverse domains \cite{hafner2025mastering}. Recent JEPA-style methods instead learn decoder-free predictive
representations, often from images or videos, by predicting target embeddings rather than reconstructing pixels
\cite{lecun2022path,assran2023self,bardes2024revisiting,mo2024connecting, assran2025v,maes2026leworldmodel}. World-model learning from large-scale videos and pretrained video models has been explored in \cite{cui2023universal} and \cite{huang2026vidworld}, respectively.
Although pixel-based world models are effective in visual domains, their representation-learning challenges differ from those arising with vector-valued observations \cite{scannell2025discrete}. The Dreamer family \cite{hafner2021mastering,hafner2025mastering} also supports vector-valued inputs, but its reported limitations on vector-based continuous-control tasks motivate WMs tailored to this setting \cite{scannell2025discrete}. 

\paragraph{Neural SSMs.}
Recurrent neural networks (RNNs), including LSTMs \cite{hochreiter1997long} and GRUs \cite{cho2014properties}, encode observation history in hidden states for sequence prediction. Kalman-style models, including Backprop KF \cite{haarnoja2016backprop}, RKN \cite{becker2019recurrent}, Ac-RKN \cite{shaj2020action}, and HiP-RSSM \cite{shaj2022hidden}, incorporate Gaussian filtering updates into neural architectures for latent-state estimation and dynamics prediction. Their evaluations primarily emphasize estimation and prediction, with less attention to task-level control and downstream decision-making. Variational autoencoder (VAE)-based control-oriented neural SSMs such as DVBF \cite{karl2017deep} jointly learn latent representations and dynamics, while NewtonianVAE \cite{jaques2021newtonianvae} imposes explicit position–velocity structure to enable proportional control in latent space. However, partially conditioned amortized inference can introduce a conditioning gap that compromises learning of latent transition dynamics \cite{bayer2021mind}. 

\paragraph{Linear and locally-linear SSMs.}  MTS3 \cite{shaj2023multi} is a prediction-oriented WM for vector-valued observations built on hierarchical linear Gaussian SSMs. Its slow-time-scale model uses aggregated action abstractions, which could complicate the translation of abstract plans into primitive actions. Control-oriented locally linear SSMs are developed in \cite{watter2015embed,banijamali2018robust}, while \cite{Levine2020Prediction} learns nonlinear latent transitions with curvature regularization to support locally linear control. However, local linear approximations may lose accuracy away from nominal trajectories and require repeated linearization during trajectory optimization.

\begin{wrapfigure}{r}{0.4\linewidth}
    \centering
    \includegraphics[width=\linewidth]{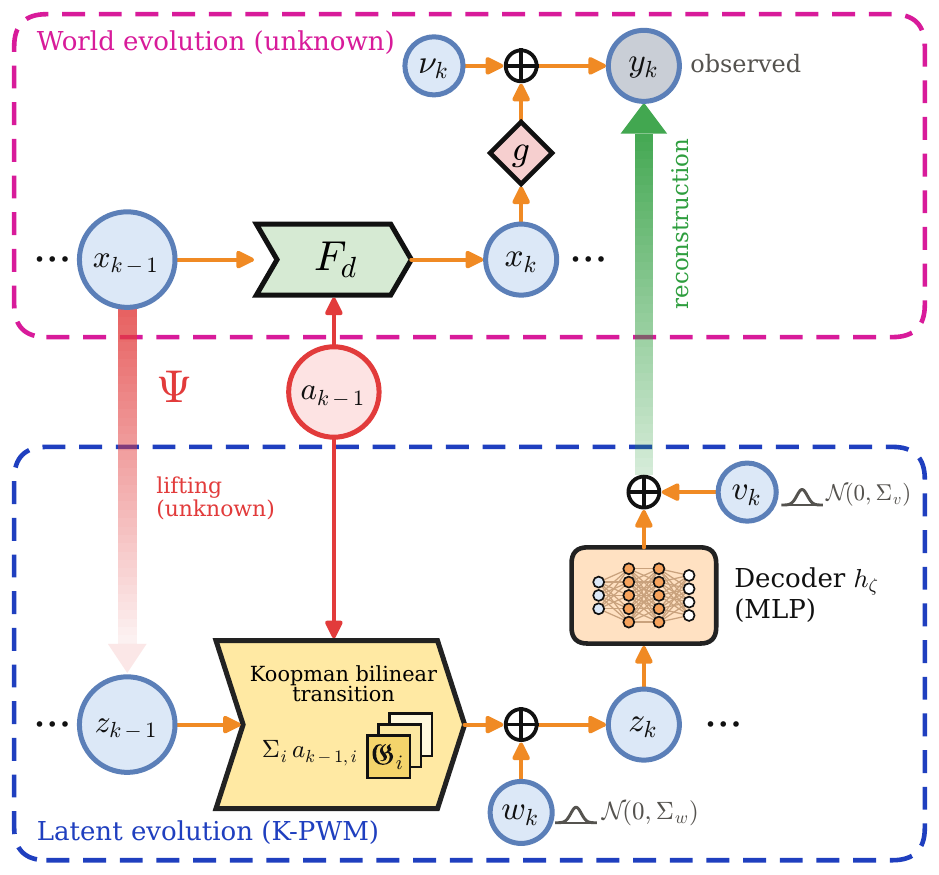}
    \caption{World and latent evolution (with \kpwm{}).}
    \label{fig:koop_view}
\end{wrapfigure}

\paragraph{Decoder-free WMs.} Decoder-free WMs such as TD-MPC \cite{hansen2022temporal}, TD-MPC2 \cite{hansen2024tdmpc}, and DCWM \cite{scannell2025discrete} learn models of latent transitions, rewards, and values without reconstructing observations, enabling online planning for continuous control. Their standard formulations do not explicitly maintain beliefs over hidden states, which may limit performance when observations or stacked histories provide insufficient state information. Moreover, online planning with densely connected neural models can incur substantial computational overhead, potentially limiting real-time deployment \cite{zhang2026factdmpc}. Fac-TDMPC \cite{zhang2026factdmpc} reduces planning latency through latent factorization and decentralized planning. However, its additive reward/value decomposition can inadequately capture dependencies among joint actions, and the authors report slower learning in some highly dynamic tasks requiring coordinated control.

\paragraph{Koopman operator-based methods.} Koopman operator theory offers a promising approach to structured world modeling through linear, bilinear, or semilinear evolution in latent space, potentially extending predictive validity across broader operating regimes \cite{koopman1931hamiltonian,mezic2005spectral,korda2018linear,goswami2021bilinearization,rajkumar2026data,rajkumar2025real,iacob2024koopman}. However, dictionary selection and learning from partial observations remain important challenges \cite{strasser2026overview}. Existing approaches include linear latent dynamics with learned dictionaries, as in DKUC \cite{shi2022deep}, and bilinear dynamics with prespecified dictionaries \cite{bruder2021advantages, folkestad2021koopman}. These formulations rely on full-state measurements, which may be unavailable in real-world robotics, and manually specified embedding dimensions, without explicit uncertainty quantification. Otto et al. \cite{otto2024learning} learn bilinear latent dynamics from partially observed trajectories without prespecifying a dictionary. However, their formulation assumes linearly independent observation functions and an affine decoder, which may limit its suitability for latent representations requiring nonlinear observation mappings, such as those derived for certain aerial robots in \cite{rajkumar2025real}. Latent-dimension selection also remains empirical.

\section{\kpwm: Koopman Bilinear-Structured Probabilistic World Model}\label{sec:kpwm}

In a partially observed setting, let $x \in \mathcal{X} \subseteq \mathbb{R}^n$ denote the true but unobserved state of an embodied system, let $a \in \mathcal{A} \subseteq \mathbb{R}^m$ represent the control input (action), and let $y \in \mathcal{Y} \subseteq \R^{n_y}$ denote the observation, which provides only partial information about the underlying state. We assume uniform sampling with interval $\Delta t$, such that for all $k \in \mathbb{N}_0$, the sampling instants are given by $t_k = k \Delta t$. The resulting discrete-time SSM of the system is given by
\alnn{
x_{k+1} = F_d(x_k, a_k), \qquad
y_k = g(x_k) + \nu_k, \quad \nu_k \sim \mathcal{N}(0, \Sigma_\nu),
\label{eq:ssm}
}
where $F_d \! : \! \mathcal{X} \! \times \! \mathcal{A} \! \to \! \mathcal{X}$ denotes the unknown discrete-time flow map induced by the underlying continuous-time dynamics
$
\dot x(t) = F\paren{ x(t), a(t)},
$
and $g:\mathcal{X}\to\mathcal{Y}$ is an observation map that is not assumed to be injective. Thus, although the underlying state transition is deterministic, only noisy, partial measurements of the state are available. This setting is common in robotics and autonomous driving \cite{Platt-RSS-10, platt2016efficient, uehara2023computationally, hao2025neural}.

To obtain a structured latent dynamics model, we assume the existence of a smooth, possibly nonlinear lifting map $\Psi=[\psi_1,\ldots,\psi_N]^\top:\mathcal{X}\to\mathcal{Z}\subseteq\mathbb{R}^N$, with $N\gg n$, and define the latent representation $z=\Psi(x)$. Here, $\mathcal{Z}=\Psi(\mathcal{X})$ is the latent state space, and each $\psi_i:\mathcal{X}\to\mathbb{R}$ is a scalar \emph{observable}, i.e., a function of the underlying state, belonging to a common function space $\mathcal{F}$, such as a Banach space. The collection $\{\psi_1,\ldots,\psi_N\}$ forms the \textit{dictionary of observables} defining the latent variables. For systems with inputs, a Koopman bilinear representation is often more suitable than a purely linear one, as it better captures control-dependent dynamics while retaining computational tractability \cite{bruder2021advantages}.

Selecting or learning a dictionary of observables that yields an accurate finite-dimensional Koopman state-space model is challenging \cite{brunton2022modern}. Partial observation compounds this challenge because the observables are defined on the underlying state $x$, whereas learning relies on noisy, partial measurements and the applied actions \cite{otto2024learning,strasser2026koopman}. Under partial observation, a finite-dimensional Koopman model generally provides only an approximation of the latent dynamics. The resulting approximation error can be modeled as process noise in the latent transition \cite{ohkubo2025koopman}. Motivated by this viewpoint, we approximate \eqref{eq:ssm} by a hidden Markov model (HMM) with a Koopman bilinear transition (KBT):
\salnn{eq:HMM}{
& \mf z_{k+1} =  \big(\mf I + \Delta t \scriptstyle {\scriptstyle \sum}_{i=0}^m a_{k_i} \mfk G_i \big)^\top \mf z_k + \mf w_k, 
\quad \mf w_k = [0 \ \ w_k^\top]^\top, \quad w_k \sim \Nc(\mf 0, \bs \Sigma_w), \label{eq:kbf} \\
& \text{and an observation (decoding) model } 
y_k = h(z_k) + v_k, \quad v_k \sim \Nc(\mf 0, \bs \Sigma_v),
}
as illustrated in \Fref{fig:koop_view}. Here, the augmented latent state is defined as $\mf z_k = [1 \ \ z_k^\top]^\top \in \mathbb{R}^{N+1}$, which allows the model to capture affine and bilinear control effects in a unified form.
The coefficient $a_{k_i}$ denotes the $i$th element of the augmented action vector $\mf a_k = [1 \ \ a_k^\top]^\top \in \mathbb{R}^{m+1}$. The matrices $\mfk G_i = [\mf 0 \ \ \mf V_i]$ are finite-dimensional approximations of Koopman generators, with the first column fixed at zero. The function $h(\cdot)$ denotes a nonlinear decoder, while $\bs\Sigma_w$ and $\bs\Sigma_v$ are the process and observation noise covariances, respectively. For notational convenience, define the action-dependent matrix approximation of the Koopman operator \vspace{-2mm}
\alnn{
\bs\Kc^a := \Big(\mf I + \Delta t\  {\scriptstyle \sum}_{i=0}^m a_{k_i}\mfk G_i \Big)
= \bmat{
  {\scriptstyle 1}     & {\scriptstyle \mf b_k^\top} \\
  {\scriptstyle \mf 0} & {\scriptstyle \mf A_k^\top}
}.
}
Then the latent KBT in the non-augmented coordinates can be written equivalently as \vspace{-2mm}
\alnn{
z_{k+1} = \mf A_k z_k + \mf b_k + w_k.
\label{eq:alt_bil}
}
We model the decoder $h(\cdot)$ as a multilayer perceptron (MLP) parameterized by $\bs\zeta$ (hereafter used as $h_{\bs \zeta}$), which collects all learnable weights and biases. The initial latent state is assigned a Gaussian prior
$
z_0 \sim \Nc(\mu_0,\bs\Sigma_0).
$
Accordingly, the HMM in \eqref{eq:HMM} is parameterized by \vspace{-2mm}
\alnn{
\Theta := \{\widetilde \Theta, \bs \zeta \}, \ \ \widetilde \Theta := \set{\mf V_0,\ldots,\mf V_m,\bs\Sigma_w,\bs\Sigma_v,\bs\Sigma_0,\mu_0}.
}
Importantly, we do not assume access to the lifting map $\Psi$, nor do we explicitly attempt to identify the dictionary of observables. Instead, we posit the HMM in \eqref{eq:HMM} as a structured latent-variable model and estimate the parameter set $\Theta$ by maximizing the likelihood of the observed data. Thus, the learned latent representation is driven by statistical consistency with the observations, rather than by explicit identification of a predefined lifting transformation.

\textbf{\kpwm{} components.} \kpwm\ has three components parameterized by \(\Theta\):
\salnn{eq:WM}{
\text{Representation / inference model:}\quad
& q_\Theta(z_k \mid y_{\le k}, a_{<k})
= \mathcal N(\mu_{k|k},\bs\Sigma_{k|k}),
\label{eq:rep_model}\\
\text{Transition model:}\quad
& p_\Theta(z_{k+1}\mid z_k,a_k)
= \mathcal N(\mf A_k z_k+\mf b_k,\bs\Sigma_w),
\label{eq:trans_model}\\
\text{Observation model:}\quad
& p_\Theta(y_k\mid z_k)
= \mathcal N(h_\Theta(z_k),\bs\Sigma_v).
\label{eq:rec_model}
}
Here, \(q_\Theta\) is not an independently learned encoder, but an extended Kalman filter (EKF)-induced
Gaussian approximation to the filtering posterior obtained from the transition
and observation models. Its dependence on \(\Theta\) arises through the
KBT parameters, the decoder, and the process and observation noise covariances.

\subsection{GECM Method for Learning the \kpwm\ Parameters}

We learn the parameters of \kpwm\ from a collection of
action--observation trajectories using an iterative expectation-maximization (EM) procedure. Given an observation sequence \(y_{0:K}\) and
the corresponding action sequence \(a_{0:K-1}\), the objective is to maximize
the conditional observed-data likelihood \(p_\Theta(y_{0:K}\mid a_{0:K-1})\),
or equivalently its log likelihood. The pair \((z_{0:K},y_{0:K})\), conditioned
on \(a_{0:K-1}\), defines the complete data for the HMM structure. Under the
Markov structure of the latent dynamics and the conditional independence of the
observations given the latent states, the joint probability density factorizes as \vspace{-3mm}
\begin{align}
p_\Theta(z_{0:K},y_{0:K}\mid a_{0:K-1})
&=
p_\Theta(z_0)
\prod_{k=0}^{K-1}
p_\Theta(z_{k+1}\mid z_k,a_k)
\prod_{k=0}^{K}
p_\Theta(y_k\mid z_k).
\label{eq:joint_density}
\end{align}
For a dataset of \(R\) independent trajectories
$
\big\{
y_{0:K}^{(r)},a_{0:K-1}^{(r)}
\big\}_{r=1}^{R},
$
the observed-data log likelihood is
\alnn{\label{eq:LL}
\mathcal{L}(\Theta)
=
{\scriptstyle \sum}_{r=1}^{R}
\log p_\Theta\!\left(
y_{0:K}^{(r)}
\,\middle|\,
a_{0:K-1}^{(r)}
\right)  
=
{\scriptstyle \sum}_{r=1}^{R}
\log
\int
p_\Theta\!\left(
z_{0:K}^{(r)},y_{0:K}^{(r)}
\,\middle|\,
a_{0:K-1}^{(r)}
\right)
\dx z_{0:K}^{(r)}.
}

Direct maximization of \eqref{eq:LL} is generally intractable, as the
observed-data likelihood requires evaluating the integral over the full latent
trajectory in \eqref{eq:LL} \cite{neal1998view,otto2024learning}, with
additional complications arising from the nonlinear observation model. To obtain a tractable
training objective, we introduce an auxiliary \emph{inference} distribution
\(q^{(r)}(z^{(r)}_{0:K})\) over latent trajectories. For notational convenience, let $Z := z_{0:K}, Y: = y_{0:K}$, $Q:=\{ q^{(r)} \}_{r=1}^R$, and $\mcr A  := a_{0:K-1}$. Then \eqref{eq:LL} admits the variational decomposition \cite{dempster1977maximum, neal1998view}
\alnn{
\mc L(\Theta) = \mcr F(\Theta, Q) + {\scriptstyle \sum}_{r=1}^R \mcr D \paren{q^{(r)}(Z^{(r)})\ || \  p_\Theta\big(Z^{(r)}\  \mid \ Y^{(r)}, \mcr A^{(r)}\big)},
\label{eq:LL_decomp}
}
where $\mcr F(\Theta, Q)$ is the variational free energy defined as
\alnn{
\mcr F(\Theta,Q) = {\scriptstyle \sum}_{r=1}^R \ \ \E_{q^{(r)}} \brak{ \log p_\Theta \big( Z^{(r)}, Y^{(r)} \ \mid \ \mcr A^{(r)} \big) } - \E_{q^{(r)}}  \brak{ \log q^{(r)}(Z^{(r)}) },
\label{eq:ELBO1}
}
and $\mcr D$ is the non-negative KL divergence between $q^{(r)}(Z^{(r)})$ and $p_\Theta\big(Z^{(r)}\  \mid \ Y^{(r)}, \mcr A^{(r)}\big)$. Therefore, $\mcr F(\Theta,Q)$ is the evidence lower bound (ELBO) for the log-likelihood, i.e., $\mc L(\Theta) \ge \mcr F(\Theta,Q)$. Moreover, if the auxiliary distribution coincides with the exact posterior for
each trajectory, i.e.,
$
q^{(r)}(Z^{(r)})
=
p_\Theta\!\left(
Z^{(r)} \mid Y^{(r)}, \mcr A^{(r)}
\right),
$
for all $r \in[1:R]$, 
then the KL term in \eqref{eq:LL_decomp} vanishes and the bound becomes tight, i.e., 
$
\mathcal L(\Theta)=\mcr F(\Theta,Q).
$ 
Starting from an initial parameter \(\Theta_0\), generalized EM approach
iteratively updates the auxiliary distribution and model parameters
\cite{neal1998view}. In the GECM formulation, the M-step is further decomposed
into conditional updates over parameter blocks \cite{meng1993maximum}. The steps involved in the GECM procedure are given below:
\begin{itemize}
    \item \textbf{(Expectation) E-step:} Update \(Q_j\) such that
    $\mathcal L(\Theta_{j-1})
    =\mcr F(\Theta_{j-1},Q_j).$
    \item \textbf{(Conditional Maximization) CM-step 1:} With \(\bs \zeta\) fixed at \(\bs \zeta_{j-1}\),
    update \(\widetilde{\Theta}_{j-1}\) to \(\widetilde{\Theta}_j\) such that
    $\mcr F\big(\{\widetilde{\Theta}_j,\bs \zeta_{j-1}\},Q_j\big)
    \ge
    \mcr F\big(\{\widetilde{\Theta}_{j-1},\zeta_{j-1}\},Q_j\big).
    $
    \item \textbf{(Conditional Maximization) CM-step 2:} With \(\widetilde{\Theta}\) fixed at
    \(\widetilde{\Theta}_j\), update \(\bs \zeta_{j-1}\) to \(\bs \zeta_j\) such that
    $
    \mcr F\big((\widetilde{\Theta}_j,\zeta_j),Q_j\big)
    \ge
    \mcr F\big((\widetilde{\Theta}_j,\zeta_{j-1}),Q_j\big),
    $
    and set $\Theta_j := (\widetilde{\Theta}_j,\zeta_j)$.
\end{itemize}

Following standard practice for tractable inference computation
\cite{ghahramani1998learning,shaj2023multi}, we employ a Gaussian approximation
for latent-state inference and learning.
Specifically, for each trajectory \(r\), we approximate the intractable smoothing
posterior \(p_\Theta(Z^{(r)}\mid Y^{(r)},\mcr A^{(r)})\) by a Gaussian
inference distribution $q^{(r)}(Z^{(r)})$
computed using an extended Kalman smoother (EKS). The distribution
$Q_j:=\{ q_j^{(r)} \}_{r=1}^R$ is characterized by smoothing moments computed under the
Koopman-bilinear transition model and the nonlinear decoder. The algorithmic version of GECM is highlighted in \Cref{alg:gecm_kpwm}.

\paragraph{Koopman-consistent initialization for GECM.}
Proper initialization of the parameters \(\Theta\), particularly the generator
matrices \(\mf V_i\), and the choice of latent dimension are crucial for stable
and effective learning. This is especially important in the actuated setting,
where the latent transition parameters must capture the action-dependent
Koopman-bilinear evolution, whereas the inferred latent states must simultaneously
disentangle the effects of the initial condition and the input sequence. As observed in \cite{otto2024learning}, the performance of EM-based learning for
actuated Koopman-generator models can be sensitive to initialization, and poor
initialization may cause the learning procedure to collapse. 
This motivates the question of whether the world model can be initialized in a
Koopman-consistent manner under partial observation, while also automating the
selection of an appropriate latent dimension. We initialize the latent world model using a
delay-coordinate Koopman construction. This construction is motivated by Takens' delay-embedding theorem \cite{takens1981detecting} and the HAVOK
framework \cite{brunton2017chaos}. Let \(\tau\) denote the uniform delay-window length, measured in samples. For
each trajectory \(r\), we assume \(K>\tau\), and in practice choose
\(K\gg \tau\) so that enough delay windows are available. We define the delay co-ordinates as $\bs \xi_k^{(r)} = [ y_k^{(r)}\ \ y_{k+1}^{(r)}\ \cdots \ \ \ y_{k+\tau-1}^{(r)}  ]^\top$ and construct the trajectory-wise Hankel matrices $\mf H^{(r)}
=
\brak{
\bs \xi_0^{(r)} \ \ \bs \xi_1^{(r)} \ \cdots \ \bs \xi_{K-\tau-1}^{(r)}
},\ 
\mf H_+^{(r)}
=
\brak{
\bs \xi_1^{(r)} \ \ \bs \xi_2^{(r)} \ \cdots \ \bs \xi_{K-\tau}^{(r)}
}.$

The corresponding Hankel matrices across all trajectories are then given by
\alnn{
\mf H
&=
\brak{
\mf H^{(1)} \ \ \mf H^{(2)} \ \cdots \ \mf H^{(R)}
},
\qquad
\mf H_+
=
\brak{
\mf H_+^{(1)} \ \ \mf H_+^{(2)} \ \cdots \ \mf H_+^{(R)}
}.
}
We compute the singular value decomposition (SVD) $\mf H = \mf U \mf S \mf W^\top$, with singular values ordered as \(\sigma_1 \ge \sigma_2 \ge \cdots\). We then retain the leading
\(r_e\) left singular vectors, collected in
$\mf U_{r_e} = [\bs u_1,\ldots,\bs u_{r_e}]$ 
corresponding to the \(r_e\) largest singular values, where
\(r_e\) is selected as the smallest integer satisfying a prescribed cumulative
energy threshold,
\big(${\scriptstyle \sum}_{i=1}^{r_e}  \sigma_i^2 /{\scriptstyle \sum}_i \sigma_i^2 \big) \ge \rho$, with $\rho$ chosen as $\rho \in (0.5,1)$ . For each trajectory \(r\), the reduced delay coordinates are obtained as
\[
\bs \phi_k^{(r)} = \mf U_{r_e}^\top \bs \xi_k^{(r)}, 
\qquad
\bs \phi_{k+1}^{(r)} = \mf U_{r_e}^\top \bs \xi_{k+1}^{(r)} .
\]
Thus, the dimension of the reduced delay representation is determined directly
from the Hankel spectrum, rather than selected by manual tuning. We transform the reduced delay coordinates into a finite-dimensional feature vector using separable Legendre polynomial features. For $\bs\phi_k^{(r)}=[\phi_{k,1}^{(r)},\ldots,\phi_{k,r_e}^{(r)}]^\top \in\mathbb R^{r_e}$ and the maximum polynomial degree $d_p$, we define the transformed feature
\alnn{
\widetilde{\mf z}_k^{(r)} = \brak{1 \ \ \widetilde z_k^{(r)^\top}}^\top, \ \widetilde z_k^{(r)}
= \begin{bmatrix}
P_1(\phi_{k,1}^{(r)})\ \ldots \ P_{d_p}(\phi_{k,1}^{(r)}) \,
\ldots \,
P_1(\phi_{k,r_e}^{(r)})\ \ldots \ P_{d_p}(\phi_{k,r_e}^{(r)})
\end{bmatrix}^\top,
}
where $\widetilde{\mf z}_k^{(r)} \in \R^{N+1}$,  $N = d_p r_e$ and \(P_d(\cdot)\) denotes the Legendre polynomial of degree \(d\). Each coordinate $\phi_{k,j}^{(r)}$ is independently expanded up to degree $d_p$, and the resulting univariate polynomial features are stacked alongside a constant observable. The constant $1$ is explicitly included, so the Legendre expansion starts at $d=1$ to avoid double-counting $P_0$. The shifted feature vector \(\widetilde{\mf z}_{k+1}^{(r)}\) is constructed
analogously from \(\bs\phi_{k+1}^{(r)}\). Let \(\widehat R\) denote the number of trajectories used for initialization. Thus, the initialization is
performed using only a randomly selected subset of \(\widehat R\) trajectories,
rather than all \(R\) trajectories. To preserve the constant observable, we use
the previously defined constrained parameterization
\(\mfk G_i=\brak{\bs 0\ \ \mf V_i}\), so that only the free block
\(\mf V_i\) is estimated. We initialize the generator matrices by solving
\alnn{
\min_{\{\mf V_i\}_{i=0}^{m}}
{\scriptstyle \sum}_{r=1}^{\what R}
{\scriptstyle \sum}_{k=0}^{K-\tau-1}
\Big\|
\tfrac{\widetilde{ z}_{k+1}^{(r)}-\widetilde{z}_{k}^{(r)}}{\Delta t}
-
\Big[
\big(
{\scriptstyle \sum}_{i=0}^{m} a_{k_i}^{(r)}\mf V_i
\big)^\top
\widetilde{\mf z}_k^{(r)}
\Big]
\Big\|_2^2
+
\lambda_g {\scriptstyle \sum}_{i=0}^{m}\|\mf V_i\|_F^2,
}
where $\lambda_g >0$ is a small regularization parameter. After obtaining the initialized generator matrices, we compute the one-step
prediction residuals on the initialization trajectories. Specifically, for each
selected trajectory \(r=1,\ldots,\widehat R\) and valid delay index
\(k=0,\ldots,K-\tau-1\), we define
$
\widehat{\widetilde z}_{k+1}^{(r)}
=
\widetilde z_k^{(r)}
+
\Delta t
\left(
{\scriptstyle \sum}_{i=0}^{m} a_{k_i}^{(r)}\mf V_i
\right)^\top
\widetilde{\mf z}_k^{(r)} .
$
The process-noise covariance $\bs \Sigma_w$, and the initial latent prior $(\mu_0, \bs \Sigma_0)$ are initialized as \vspace{-3mm}
\alnn{
\bs\Sigma_w
=
\operatorname{Cov}
\big(
\widetilde z_{k+1}^{(r)}
-
\widehat{\widetilde z}_{k+1}^{(r)}
\big)_{r,k},\quad
\mu_0
=
\tfrac{1}{\widehat R}
{\scriptstyle \sum}_{r=1}^{\widehat R}
\widetilde z_0^{(r)},
\quad
\bs\Sigma_0
=
\operatorname{Cov}
\left(
\widetilde z_0^{(r)}
\right)_{r=1}^{\widehat R}.
}
Thus, \(\bs\Sigma_w\), \(\mu_0\), and \(\bs\Sigma_0\) are initialized directly
from the residual statistics and initial transformed features induced by the
Koopman-consistent initialization. 

\paragraph{CM-Step 1.}
In CM-step 1 of EM iteration \(j\), given the inference distribution \(Q_j\)
and with \(\bs\zeta\) fixed at \(\bs\zeta_{j-1}\), we update
\(\widetilde{\Theta}_{j-1}\) by maximizing the ELBO \eqref{eq:ELBO1}. The required moments under \(q_j^{(r)}\) are denoted by
$
\mu_{k}^{(r)}
=
\mathbb E_{q_j^{(r)}}\!\big[z_k^{(r)}\big]$, 
$\bs\Sigma_{k,k}^{(r)}
=
\mathbb E_{q_j^{(r)}}\!\Big[
\big( z_k^{(r)}-\mu_k^{(r)}\big)
\big( z_k^{(r)}-\mu_k^{(r)}\big)^\top
\Big],
$
and
$
\bs\Sigma_{k,k+1}^{(r)}
=
\mathbb E_{q_j^{(r)}}\!\Big[
\big( z_k^{(r)}-\mu_k^{(r)}\big)
\left( z_{k+1}^{(r)}-\mu_{k+1}^{(r)}\right)^\top
\Big].
$
Recalling that the augmented latent state is $\mf z_k  = [1 \ \ z_k^{(r)^\top}]^\top$,  let $\dot{z}_k^{(r)}
= \tfrac{ z_{k+1}^{(r)}- z_k^{(r)}}{\Delta t}$. Let \(\mc M_k^{(r)}\) denote the second-moment matrix of the augmented latent state, and let \(\mc D_k^{(r)}\) denote the cross-moment matrix between the latent-state derivative and the augmented latent state; these matrices are given by $
\mc M_k^{(r)}
:=
\mathbb E_{q_j^{(r)}}
\!\big[
\mf z_k^{(r)}\mf z_k^{(r)\top}
\big]$, and
$
\mc D_k^{(r)}
:=
\mathbb E_{q_j^{(r)}}
\!\big[
\dot z_k^{(r)}\mf z_k^{(r)\top}
\big].
$
Let
\alnn{
\mf M
=
{\scriptstyle \sum}_{r=1}^{R}
{\scriptstyle \sum}_{k=0}^{K-1}
\big(
\bs a_k^{(r)}\bs a_k^{(r)\top}
\big)
\otimes
\mc M_k^{(r)}, \text{ and } \mf D
=
{\scriptstyle \sum}_{r=1}^{R}
{\scriptstyle \sum}_{k=0}^{K-1}
\bs a_k^{(r)\top}
\otimes
\mc D_k^{(r)}.
}
Now we state the key result for the CM-step 1.
\begin{assumption}\label{assum:ps}
The matrices $\bs\Sigma_w$, $\bs\Sigma_v$, and $\bs\Sigma_0$ remain positive definite, and the
training action sequences are persistently exciting (see \Cref{sub:assumpe}).
\end{assumption}

\begin{proposition}\label{prop:1}
Under \Aref{assum:ps}, CM-step~1 yields the closed-form ELBO maximizer
$\widetilde{\Theta}_j$ with the decoder parameters $\bs\zeta_{j-1}$
held fixed. The complete statement of Proposition~\ref{prop:1}
 and its proof appear in \Cref{prop1:statement} and \Cref{proof:prop1}, respectively.
\end{proposition}

\begin{remark}
In high-dimensional settings, EM updates can suffer from numerical instability and ill-conditioned estimates~\cite{yi2015regularized}. Accordingly, \Aref{assum:ps} can be enforced in CM-step 1 by adding small positive regularization terms to the closed-form updates, equivalent to maximizing the regularized ELBO $\widehat{\mcr F}(\Theta,Q):=\mcr F(\Theta,Q)-\mcr R(\widetilde{\Theta})$, where $\mcr R(\widetilde{\Theta})$ penalizes ill-conditioned Koopman-bilinear transition parameters and covariance estimates.
\end{remark}

\paragraph{CM-Step 2.}

In CM-step 2 of EM iteration \(j\), given \(Q_j\) and with \(\widetilde{\Theta}\) fixed at \(\widetilde{\Theta}_j\), we update the decoder parameters from \(\bs\zeta_{j-1}\) to \(\bs\zeta_j\). We summarize the key result for CM-Step 2 below.

\begin{proposition}[Decoder update increases the CM-step objective]
\label{prop:2}
Fix  \(Q_j\) and the parameters
\(\widetilde{\Theta}_j\), and define the decoder objective
$
\mc L_{\mathrm{dec}}(\bs\zeta)
:=
\tfrac{1}{R(K+1)}
{\scriptstyle \sum}_{r=1}^{R}
{\scriptstyle \sum}_{k=0}^{K}\ 
\E_{q_j^{(r)}}
\Big[
\big\|
y_k^{(r)}-h_{\bs\zeta}(z_k^{(r)})
\big\|_{\bs\Sigma_{v,j}^{-1}}^2
\Big].
$
If the decoder parameter update from \(\bs\zeta_{j-1}\) to \(\bs\zeta_j\) satisfies
$
\mc L_{\mathrm{dec}}(\bs\zeta_j)
\le
\mc L_{\mathrm{dec}}(\bs\zeta_{j-1}),
$
then
$
\mcr F\big((\widetilde{\Theta}_j,\bs\zeta_j),Q_j\big)
\ge
\mcr F\big((\widetilde{\Theta}_j,\bs\zeta_{j-1}),Q_j\big).
$
Consequently, such an update is a valid generalized CM-step 2 update. We provide the proof in \Sref{proof:prop2}.
\end{proposition}

\paragraph{E-Step.}
During the E-step of the EM iteration \(j\), for each trajectory \(r\), we compute the Gaussian
smoothing approximation
\[
q_j^{(r)}\!\big(Z^{(r)}\big)
=
\mathcal N\!\big(Z^{(r)};\mu_{0:K}^{(r)},\bs\Sigma_{0:K}^{(r)}\big)
\approx
p_{\Theta_{j-1}}\!\big(
Z^{(r)}\mid Y^{(r)},\mathcal A^{(r)}
\big)
\]
using an EKF followed by a Rauch--Tung--Striebel backward
smoother \cite{sarkka2023bayesian, rauch1965maximum}. For brevity, we refer to this procedure as extended Kalman smoothing (EKS). For notational simplicity, we suppress the trajectory index \(r\)
below. We now explain the EKF and smoothing process under E-step below.

\paragraph{Forward pass (EKF).} Given the Koopman-bilinear transition \eqref{eq:HMM} and \eqref{eq:alt_bil}, the forward EKF prediction step is given by \cite[Section~13.2.3]{simon2006optimal}, \cite{sarkka2023bayesian} 
\alnn{
\mu_{k+1|k}
=
\mf A_{k,j-1} \mu_{k|k}
+
\bs b_{k,j-1}, \quad 
\bs\Sigma_{k+1|k}
=
\mf A_{k,j-1}\bs\Sigma_{k|k}\mf A_{k,j-1}^{\top}
+
\bs\Sigma_{w,j-1}.}
The decoder linearized at the predicted mean:
$
\mf J_{k+1,j-1}
:=
\nabla_{ z}
h_{\bs\zeta_{j-1}}( z)
\big|_{ z=\mu_{k+1|k}} .
$
Then the EKF update is given by
\alnn{
 \bs S_{k+1}
 & =
\mf J_{k+1,j-1}\bs\Sigma_{k+1|k}\mf J_{k+1,j-1}^{\top}
+
\bs\Sigma_{v,j-1}, \quad 
\mf L_{k+1}
=
\bs\Sigma_{k+1|k}\mf J_{k+1,j-1}^{\top}\bs S_{k+1}^{-1}, \\
 \mu_{k+1|k+1}
&=
\mu_{k+1|k}
+
\mf L_{k+1}
\big(
 y_{k+1}
-
h_{\bs\zeta_{j-1}}(\mu_{k+1|k})
\big),\\
 \bs\Sigma_{k+1|k+1}
&=
\big(
\mf I-\mf L_{k+1}\mf J_{k+1,j-1}
\big)
\bs\Sigma_{k+1|k},
}
where $\mf L_{k+1}$ is the Kalman gain.

\paragraph{Backward pass (smoothing)} After the forward pass, the backward smoother is initialized by
$
\mu_{K}=\mu_{K|K},\ 
\bs\Sigma_{K,K}=\bs\Sigma_{K|K}.
$
For \(k=K-1,\ldots,0\), define the smoother gain as
$
\mf C_k
=
\bs\Sigma_{k|k}\mf A_{k,j-1}^{\top}
\bs\Sigma_{k+1|k}^{-1}.
$
The smoothed moments are then computed as \cite{sarkka2023bayesian}
\alnn{
\mu_k
=
\mu_{k|k}
+
\mf C_k
\left(
\mu_{k+1}
-
\mu_{k+1|k}
\right), \ 
\bs\Sigma_{k,k}
=
\bs\Sigma_{k|k}
+
\mf C_k
\left(
\bs\Sigma_{k+1,k+1}
-
\bs\Sigma_{k+1|k}
\right)
\mf C_k^{\top}.
}
The lag-one smoothing covariance used in the CM-step is approximated by
$
\bs\Sigma_{k,k+1}
=
\mf C_k\bs\Sigma_{k+1,k+1}$, with
$\bs\Sigma_{k+1,k}
=
\bs\Sigma_{k,k+1}^{\top}.
$
Thus, for each trajectory, the EKS provides the moments which are used in the subsequent CM-step updates.

\section{Uncertainty-aware Model Predictive Path Integral (MPPI) Control} \label{sec:MPPI}

We use MPPI as a sampling-based receding-horizon controller
\cite{williams2017mppi,williams2018information}, building on path-integral
stochastic optimal control \cite{theodorou2010generalized}. To make our planning
aware of predictive uncertainty, we augment the MPPI roll-out objective with a
covariance-dependent risk penalty, in the spirit of risk-aware MPPI variants
\cite{yin2023risk}. After learning K-PWM, we use the
learned latent SSM as the dynamics model inside MPPI. At each environment step,
the current observation is assimilated using the EKF induced by the learned
transition and decoder models, producing a filtering belief
$(\mu_{k|k},\bs\Sigma_{k|k})$ initialized from the latent prior
$(\mu_0,\bs\Sigma_0)$. The MPPI scheme then samples $S$ candidate action sequences around
a shifted nominal sequence and rolls out the Koopman-bilinear latent dynamics
for a horizon of $L$ steps from this belief. During each roll-out, the predicted latent
mean is decoded to observation space for task-cost evaluation, while the latent
covariance is propagated to estimate model uncertainty along the trajectory. Further details on the MPPI scheme and its implementation are provided in \Sref{sub:mppi_supp} and \Cref{alg:mppi_kpwm}.

\section{Experimental Evaluations} \label{sec:exp}

We evaluate the data efficiency of \kpwm\ in two sets of experiments focused on prediction and continuous control. In the prediction experiments, we measure prediction accuracy under partial observability and with limited training data on selected Gymnasium MuJoCo systems \cite{towers2026gymnasium}. In the control experiments, we assess whether models learned from limited data support effective continuous control on high-dimensional tasks built using robot models from MuJoCo Menagerie \cite{menagerie2022github}. To substantiate the data efficiency of \kpwm\, we limit the training data to 15 ($R=15$) independently sampled action--observation trajectories per task across all experiments. 

\paragraph{Prediction under partial-observation.} We compare the prediction performance of \kpwm{} against five prediction-oriented baselines on three partially observed Gymnasium MuJoCo systems (Half Cheetah, Hopper, and Swimmer). The baselines are MTS3~\cite{shaj2023multi}, LSTM~\cite{hochreiter1997long}, GRU~\cite{cho2014properties}, HiP-RSSM~\cite{shaj2022hidden}, and Ac-RKN~\cite{shaj2020action}. For each system, we independently generate training trajectories by sampling random initial states and applying piecewise-constant random actions within system-specific bounds, with observations corrupted by additive Gaussian measurement noise. To introduce a strict partial observation scenario, we retain 9 of 17 observation components for Half Cheetah, 7 of 11 for Hopper, and 7 of 10 for Swimmer, masking the remaining 8, 4, and 3 components, respectively, in the training data.
\begin{table*}[htpb!]
\centering
\begin{minipage}[t]{0.54\textwidth}
  \vspace{0pt}
  \centering
  \caption{Prediction on Gymnasium MuJoCo systems: RMSE over a prediction horizon $T_H$ ($\mrm s$).}
  \label{tab:gym_rmse}
  \scriptsize
  \setlength{\tabcolsep}{3.5pt}
  \begin{tabular}{lccc}
    \toprule
    Method & Hopper & Swimmer & Half Cheetah \\
    \midrule
    K-PWM
    & \heatcell{5}{\mathbf{0.0087 \pm 0.0028}}
    & \heatcell{5}{\mathbf{0.093 \pm 0.015}}
    & \heatcell{5}{\mathbf{0.048 \pm 0.0060}} \\
    MTS3
    & \heatcell{10}{0.014 \pm 0.0027}
    & \heatcell{21}{0.18 \pm 0.025}
    & \heatcell{28}{0.14 \pm 0.018} \\
    LSTM
    & \heatcell{39}{0.045 \pm 0.012}
    & \heatcell{100}{0.61 \pm 0.11}
    & \heatcell{80}{0.35 \pm 0.065} \\
    GRU
    & \heatcell{39}{0.045 \pm 0.016}
    & \heatcell{76}{0.48 \pm 0.051}
    & \heatcell{78}{0.34 \pm 0.13} \\
    HiP-RSSM
    & \heatcell{43}{0.049 \pm 0.011}
    & \heatcell{38}{0.27 \pm 0.028}
    & \heatcell{93}{0.40 \pm 0.13} \\
    Ac-RKN
    & \heatcell{100}{0.11 \pm 0.0064}
    & \heatcell{36}{0.26 \pm 0.032}
    & \heatcell{100}{0.43 \pm 0.12} \\
    \bottomrule
  \end{tabular}

  \par\vspace{2pt}
  \textit{Note:} $T_H$ (Hopper, Swimmer, Half Cheetah): $(0.2, 6, 12)\,\mathrm{s}$. RMSEs reported as mean $\pm$ standard deviation over 10 test trajectories.
\end{minipage}\hfill
\begin{minipage}[t]{0.46\textwidth}
  \vspace{0pt}
  \centering
  \includegraphics[width=\linewidth]{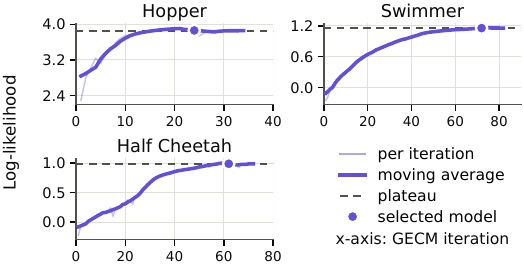}
  \captionof{figure}{\kpwm\ training log-likelihood on Gymnasium MuJoCo systems.}
  \label{fig:ll_gym}
\end{minipage}
\end{table*}
\vspace{-0.7em}

\begin{table*}[htpb!]
\centering
\begin{minipage}[t]{0.49\textwidth}
\vspace{0pt}
  \centering
  \includegraphics[width=\linewidth]{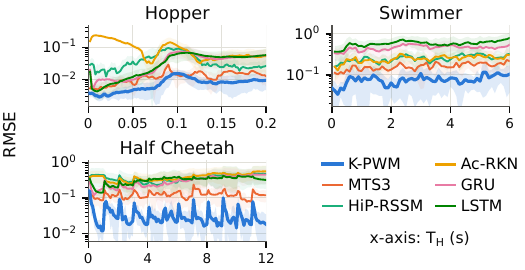}
  \captionof{figure}{ RMSEs with increasing prediction horizon on Gymnasium MuJoCo tasks.}
  \label{fig:gym_rmse_over_horizon}

\end{minipage}
\hfill
\begin{minipage}[t]{0.49\textwidth}
  \vspace{0pt}
  \centering
  \includegraphics[width=\linewidth]{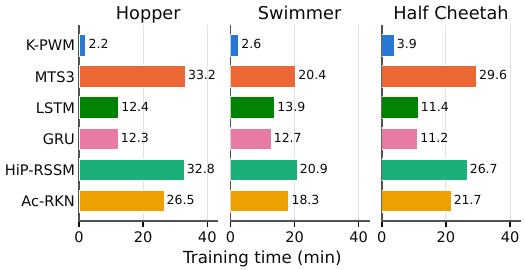}
  \captionof{figure}{Training time for Gymnasium MuJoCo tasks.}
  \label{fig:gym_training_time}
\end{minipage}

\end{table*}

\Cref{fig:ll_gym} shows the \kpwm{} training log-likelihood over GECM iterations. The moving averages plateau near iterations 24, 72, and 62 for Hopper, Swimmer, and Half Cheetah, respectively. The markers indicate the models retained at those iterations. The rise and subsequent plateau in log-likelihood show that GECM improves the fit of \kpwm{} to the training data across all three systems. \Cref{tab:gym_rmse} shows that \kpwm{} achieves the lowest mean RMSE over the prediction horizon $T_H$ for each system. Compared with MTS3, the most accurate baseline, \kpwm{} reduces mean RMSE by approximately 38\%, 48\%, and 66\% on Hopper, Swimmer, and Half Cheetah, respectively. \kpwm{} also has the shortest training time on each system with a $7.6$--$15.1\times$ training speedup over MTS3, and \kpwm{} has the shortest training time across all three benchmarks as presented in \Cref{fig:gym_training_time}.
\Cref{fig:gym_rmse_over_horizon} shows mean prediction RMSE as a function of $T_H$. Across the tested horizons, \kpwm{} generally achieves the lowest RMSE, with a clear advantage on Half Cheetah and Swimmer and performance close to MTS3 on Hopper near $T_H = 0.1\,\mathrm{s}$.

\paragraph{Continuous-control in robotic systems.}
We next evaluate the effectiveness of \kpwm{} for continuous control of realistic robotic systems when trained on limited data, along with its potential for real-time deployment. We consider four high-dimensional robotic systems from the MuJoCo Menagerie suite: the Franka Emika Panda, TIAGo++, ALOHA 2, and Allegro Hand. We compare against E2C~\cite{watter2015embed}, RCE~\cite{banijamali2018robust}, DeepKoopman (DK)~\cite{shi2022deep}, and the best-performing prediction-oriented baseline in the preceding experiments (MTS3). For each robot, we independently generate training trajectories by sampling random initial configurations and applying filtered random joint-position commands. We then evaluate the trained models on a circular trajectory-tracking task for each robot, using its end effector (EE) or multiple EEs, as applicable. All methods are evaluated using the presented MPPI scheme with an identical controller configuration, making minor method-specific adaptations where needed to integrate each model. \Cref{fig:ll_menagerie} shows that the training log-likelihood of \kpwm{} increases and approaches a plateau on all four robots, with the selected models lying near these plateaus. \Cref{fig:men_training_time} reports training times of $6.3$--$22.6$ minutes for \kpwm{}, making it $1.9$--$5.6\times$ faster than MTS3 and faster than DK on every robot, while RCE and E2C train more quickly.

\begin{table*}[htpb!]
\centering
\begin{minipage}[t]{0.45\textwidth}
\vspace{0pt}
  \centering
  \includegraphics[width=\linewidth]{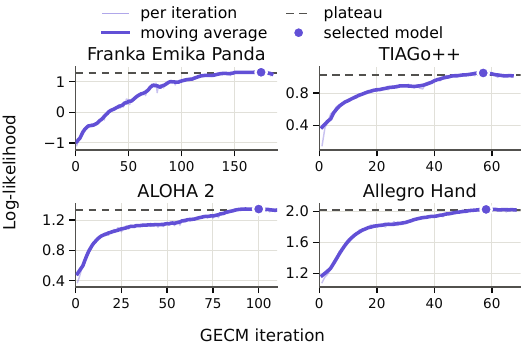}
  \captionof{figure}{\kpwm{} training log-likelihood on MuJoCo Menagerie systems.}
  \label{fig:ll_menagerie}

\end{minipage}
\hfill
\begin{minipage}[t]{0.45\textwidth}
  \vspace{0pt}
  \centering
  \includegraphics[width=\linewidth]{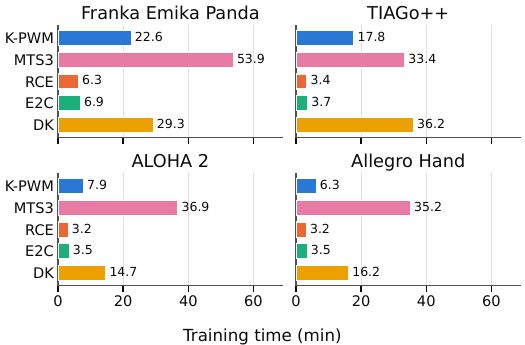}
  \captionof{figure}{Training time for MuJoCo Menagerie systems.}
  \label{fig:men_training_time}
\end{minipage}

\end{table*}

\begin{table*}[htpb!]
\centering
\begin{minipage}[t]{0.49\textwidth}
  \vspace{0pt}
  \centering
  \caption{EE tracking performance with MPPI controller: EE tracking RMSE ($\mrm{mm})$.}
  \label{tab:ee_track_rmse}
  \scriptsize
  \setlength{\tabcolsep}{3pt}
  \begin{tabular}{lcccc}
    \toprule
    Method & \shortstack{Franka Emika\\Panda}
           & TIAGo++ & ALOHA 2 & \shortstack{Allegro\\Hand} \\
    \midrule
    K-PWM
    & \heatcell{10}{6.1 \pm 0.059}
    & \heatcell{5}{\mathbf{1.9 \pm 0.236}}
    & \heatcell{49}{2.5 \pm 0.037}
    & \heatcell{5}{\mathbf{0.8 \pm 0.002}} \\
    MTS3
    & \heatcell{30}{6.6 \pm 0.028}
    & \heatcell{41}{2.6 \pm 0.138}
    & \heatcell{92}{2.9 \pm 0.032}
    & \heatcell{15}{0.9 \pm 0.006} \\
    RCE
    & \heatcell{5}{\mathbf{6.0 \pm 0.032}}
    & \heatcell{21}{2.2 \pm 0.103}
    & \heatcell{94}{3.0 \pm 0.044}
    & \heatcell{12}{0.9 \pm 0.002} \\
    E2C
    & \heatcell{100}{8.5 \pm 0.027}
    & \heatcell{21}{2.2 \pm 0.229}
    & \heatcell{100}{3.0 \pm 0.031}
    & \heatcell{100}{1.6 \pm 0.016} \\
    DK
    & \heatcell{54}{7.3 \pm 0.037}
    & \heatcell{100}{3.7 \pm 0.062}
    & \heatcell{5}{\mathbf{2.1 \pm 0.034}}
    & \heatcell{6}{0.8 \pm 0.002} \\
    \bottomrule
  \end{tabular}

  \par\vspace{2pt}
  \textit{Note:} RMSEs are reported as mean $\pm$ standard
  deviation over 5 independent runs.
\end{minipage}\hfill
\begin{minipage}[t]{0.49\textwidth}
  \vspace{0pt}
  \centering
  \caption{Per step computation time in $\mrm{ms}$ with MPPI controller}
  \label{tab:ee_track_time}
  \scriptsize
  \setlength{\tabcolsep}{3pt}
  \begin{tabular}{lcccc}
    \toprule
    Method & \shortstack{Franka Emika\\Panda}
           & TIAGo++ & ALOHA 2 & \shortstack{Allegro\\Hand} \\
    \midrule
    K-PWM
    & \heatcell{6}{5.6 \pm 0.020}
    & \heatcell{20}{22.0 \pm 1.830}
    & \heatcell{6}{6.3 \pm 0.070}
    & \heatcell{6}{6.3 \pm 0.060} \\
    MTS3
    & \heatcell{100}{94.4 \pm 5.300}
    & \heatcell{100}{95.1 \pm 3.130}
    & \heatcell{100}{96.4 \pm 3.050}
    & \heatcell{100}{98.0 \pm 3.850} \\
    RCE
    & \heatcell{12}{11.1 \pm 0.120}
    & \heatcell{31}{31.3 \pm 1.030}
    & \heatcell{12}{12.4 \pm 0.020}
    & \heatcell{13}{13.0 \pm 0.070} \\
    E2C
    & \heatcell{10}{9.6 \pm 0.040}
    & \heatcell{25}{26.7 \pm 1.430}
    & \heatcell{11}{11.1 \pm 0.190}
    & \heatcell{12}{11.8 \pm 0.310} \\
    DK
    & \heatcell{5}{\mathbf{4.9 \pm 0.290}}
    & \heatcell{5}{\mathbf{7.9 \pm 0.120}}
    & \heatcell{5}{\mathbf{5.8 \pm 0.190}}
    & \heatcell{5}{\mathbf{5.1 \pm 0.030}} \\
    \bottomrule
  \end{tabular}

  \par\vspace{2pt}
  \textit{Note:} Computation time is reported as mean $\pm$ standard
  deviation over 5 independent runs.
\end{minipage}
\end{table*}

\Cref{tab:ee_track_rmse} reports EE tracking errors over five independent runs. \kpwm{} consistently delivers competitive accuracy, ranking among the two most accurate methods on every robot and achieving the lowest mean error on TIAGo++ and Allegro Hand. \Cref{tab:ee_track_time} shows that it is also the second-fastest method throughout, providing $1.2$--$2.1\times$ speedups over RCE and E2C and $4.3$--$17.0\times$ over MTS3. Its mean computation time falls below the $20\,\mrm{ms}$ control period on three robots. On TIAGo++, it remains competitive for real-time control at $22.0\,\mrm{ms}$ per step, exceeding the sampling period by only $2.0\,\mrm{ms}$. Although DK is faster, \kpwm{} achieves lower mean tracking errors on three robots. Together with its shorter training time than MTS3 and DK, these results demonstrate competitive tracking from limited data with efficient training and online control.

\begin{figure*}[htpb!]
    \centering
    \begin{subfigure}[t]{0.44\textwidth}
        \centering
        \includegraphics[width=\linewidth]{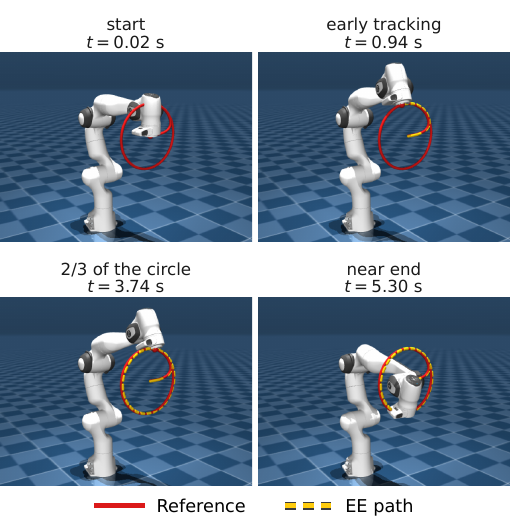}
        \caption{Franka Emika Panda on MuJoCo.}
        \label{fig:result_a}
    \end{subfigure}\hfill
    \begin{subfigure}[t]{0.44\textwidth}
        \centering
        \includegraphics[width=\linewidth]{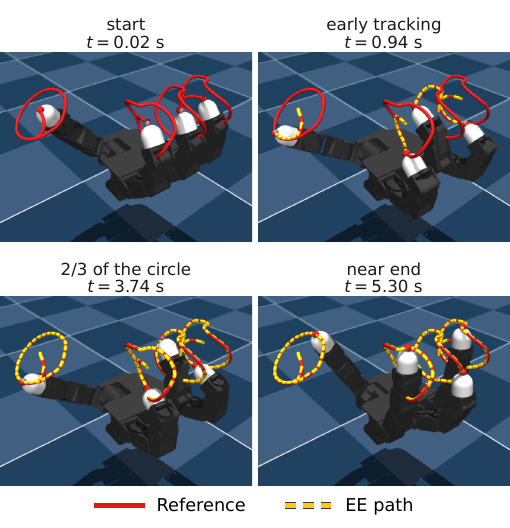}
        \caption{Allegro Hand on MuJoCo.}
        \label{fig:result_b}
    \end{subfigure}
    \caption{\kpwm{}+MPPI Control: representative EE trajectory tracking demonstration.}
    \label{fig:results}
\end{figure*}

\section{Conclusion} 
We introduced \kpwm{}, a probabilistic world model that couples Koopman-bilinear latent dynamics, nonlinear decoding, and model-induced inference for prediction and control under noisy partial observation. Its structured latent dynamics enable exact propagation of the maintained Gaussian belief through the latent transition and efficient rollouts for uncertainty-aware MPPI control. The proposed GECM procedure combines closed-form conditional updates with gradient-based decoder learning, supported by Koopman-consistent initialization and data-driven latent-dimension selection. Across partially observed Gymnasium MuJoCo systems, \kpwm{} achieves lower long-horizon prediction error than the evaluated baselines while substantially reducing training time relative to the strongest prediction baseline. On four high-dimensional MuJoCo Menagerie robots, it delivers competitive tracking accuracy with low per-step computation times, demonstrating potential for real-time deployment. These findings support structured probabilistic latent dynamics as a data- and compute-efficient foundation for prediction and control. Future work will explore richer belief representations, online adaptation, and deployment on physical robots.

\newpage

\bibliographystyle{unsrt}
\bibliography{references}

\clearpage
\appendix

\startcontents[appendices]
\printcontents[appendices]{l}{1}{%
    \section*{Appendix Contents}
    \setcounter{tocdepth}{3}%
}
\clearpage

\section*{Supplementary Materials : Appendix}

\section{Notations} \label{sub:notations}
 Let $\N:=\{1,2,\cdots\}$, $\N_0 := \{0\}\cup\N$, $\R:=(-\infty, \infty)$, $\R^+ := (0,+\infty)$, $\R^+_0 := [0,+\infty)$,  and $[a:b]:=[a,b] \cap \N_0$ for any $a,b \in \N_0$ with $b>a$. The Kronecker product is denoted by $\kron$, the gradient operator by $\nabla$, and function composition by $\circ$. For $a,\ b \in \N_0$ with $b > a$, we define $v_{a:b}:= \{ v_a,\cdots, v_{b} \}$.
The identity matrix is denoted by $\mathbf{I}$, the transpose by $(\cdot)^\top$, and the
trace and determinant by $\operatorname{tr}(\cdot)$ and $\det(\cdot)$. For a symmetric matrix
$\mathbf{A}$, $\mathbf{A}\succ 0$ denotes positive definiteness. $\|\cdot\|_2$ and
$\|\cdot\|_F$ denote the Euclidean and Frobenius norms, and
$\|x\|_{\mathbf{W}}^2 := x^\top \mathbf{W} x$ for $\mathbf{W}\succ 0$.
$\mathcal{N}(\mu,\Sigma)$ denotes a Gaussian distribution with mean $\mu$ and covariance
$\Sigma$, and $\mathcal{N}(x;\mu,\Sigma)$ its density evaluated at $x$.
$\mathbb{E}_q[\cdot]$ denotes expectation under the distribution $q$, and
$\operatorname{Cov}(\cdot)$ the sample covariance. For a sequence $\{v_k\}$, we write
$v_{\le k} := v_{0:k}$ and $v_{<k} := v_{0:k-1}$. Estimates conditioned on observations
up to time $l$ are written $\mu_{k|l}$ and $\Sigma_{k|l}$; the smoothed moments
$\mu_{k|K}$ and $\Sigma_{k|K}$ are abbreviated $\mu_k$ and $\Sigma_{k,k}$. The superscript
$(r)$ indexes trajectories, and the subscript $j$ indexes GECM iterations. Bold
$\mathbf{z}_k = [1\; z_k^\top]^\top$ and $\mathbf{a}_k = [1\; a_k^\top]^\top$ denote the
latent state and action augmented with a constant. $\dot{x}$ denotes the time derivative,
$:=$ a definition, and $\operatorname{ReLU}(x) := \max\{0, x\}$.

\begin{table}[htpb!]
\centering
\caption{Acronyms used throughout the paper.}
\label{tab:acronyms}
\scriptsize
\renewcommand{\arraystretch}{1.08}
\begin{tabular}{@{}lp{0.36\linewidth}|lp{0.36\linewidth}@{}}
\toprule
\textbf{Acronym} & \textbf{Definition} &
\textbf{Acronym} & \textbf{Definition} \\
\midrule
Ac-RKN & Action-Conditional Recurrent Kalman Network &
AI & Artificial Intelligence \\
CM & Conditional Maximization &
CPU & Central Processing Unit \\
DCWM & Discrete Codebook World Model &
DK & DeepKoopman \\
DKUC & Deep Koopman Operator with Control &
DVBF & Deep Variational Bayes Filter \\
E2C & Embed to Control &
EDMD & Extended Dynamic Mode Decomposition \\
EE & End Effector &
EKF & Extended Kalman Filter \\
EKS & Extended Kalman Smoother &
ELBO & Evidence Lower Bound \\
EM & Expectation--Maximization &
Fac-TDMPC & Factored TD-MPC \\
GECM & Generalized Expectation Conditional Maximization &
GPU & Graphics Processing Unit \\
GRU & Gated Recurrent Unit &
HAVOK & Hankel Alternative View of Koopman \\
HiP-RSSM & Hidden-Parameter Recurrent State-Space Model &
HMM & Hidden Markov Model \\
IMU & Inertial Measurement Unit &
JEPA & Joint-Embedding Predictive Architecture \\
KBT & Koopman Bilinear Transition &
KL & Kullback--Leibler \\
K-PWM & Koopman Bilinear-Structured Probabilistic World Model &
LSTM & Long Short-Term Memory \\
MBRL & Model-Based Reinforcement Learning &
MFRL & Model-Free Reinforcement Learning \\
MLP & Multilayer Perceptron &
MPC & Model Predictive Control \\
MPPI & Model Predictive Path Integral &
MTS3 & Multi-Time-Scale State-Space Model \\
MuJoCo & Multi-Joint Dynamics with Contact &
PE & Persistency of Excitation \\
RAM & Random-Access Memory &
RCE & Robust Controllable Embedding \\
ReLU & Rectified Linear Unit &
RKN & Recurrent Kalman Network \\
RL & Reinforcement Learning &
RMSE & Root Mean Squared Error \\
RNN & Recurrent Neural Network &
RTS & Rauch--Tung--Striebel \\
SSM & State-Space Model &
SVD & Singular Value Decomposition \\
TD-MPC & Temporal-Difference Model Predictive Control &
VAE & Variational Autoencoder \\
VRAM & Video Random-Access Memory &
WM & World Model \\
\bottomrule
\end{tabular}
\end{table}

\newpage 

\section{Supplementary Information for \Cref{sec:intro}}

\subsection{Inspiration behind world models.}
\begin{figure}[htpb!]
    \centering
    \begin{subfigure}[t]{0.49\linewidth}
        \centering
        \includegraphics[width=\linewidth]{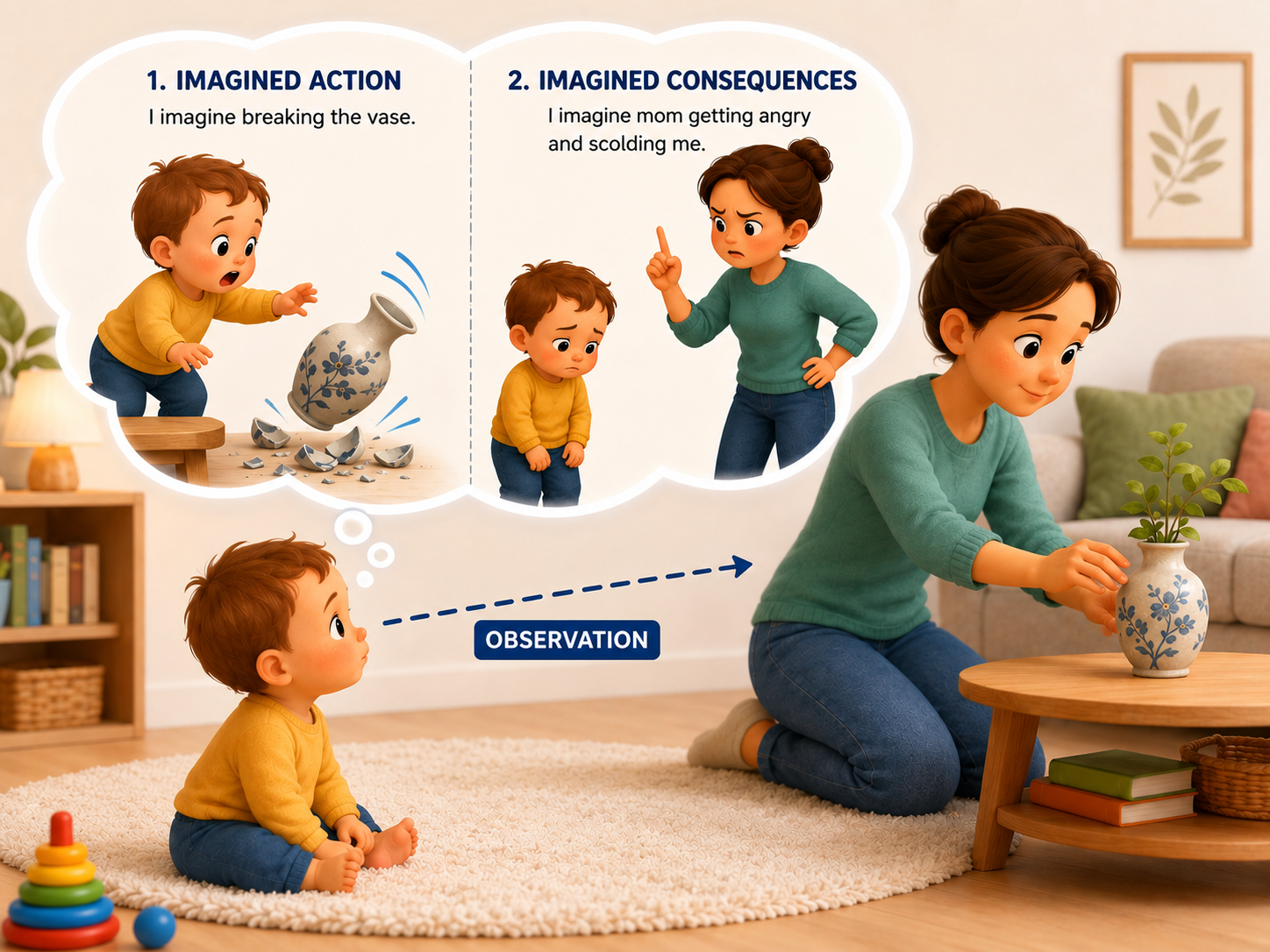}
    \end{subfigure}\hfill
    \begin{subfigure}[t]{0.49\linewidth}
        \centering
        \includegraphics[width=\linewidth]{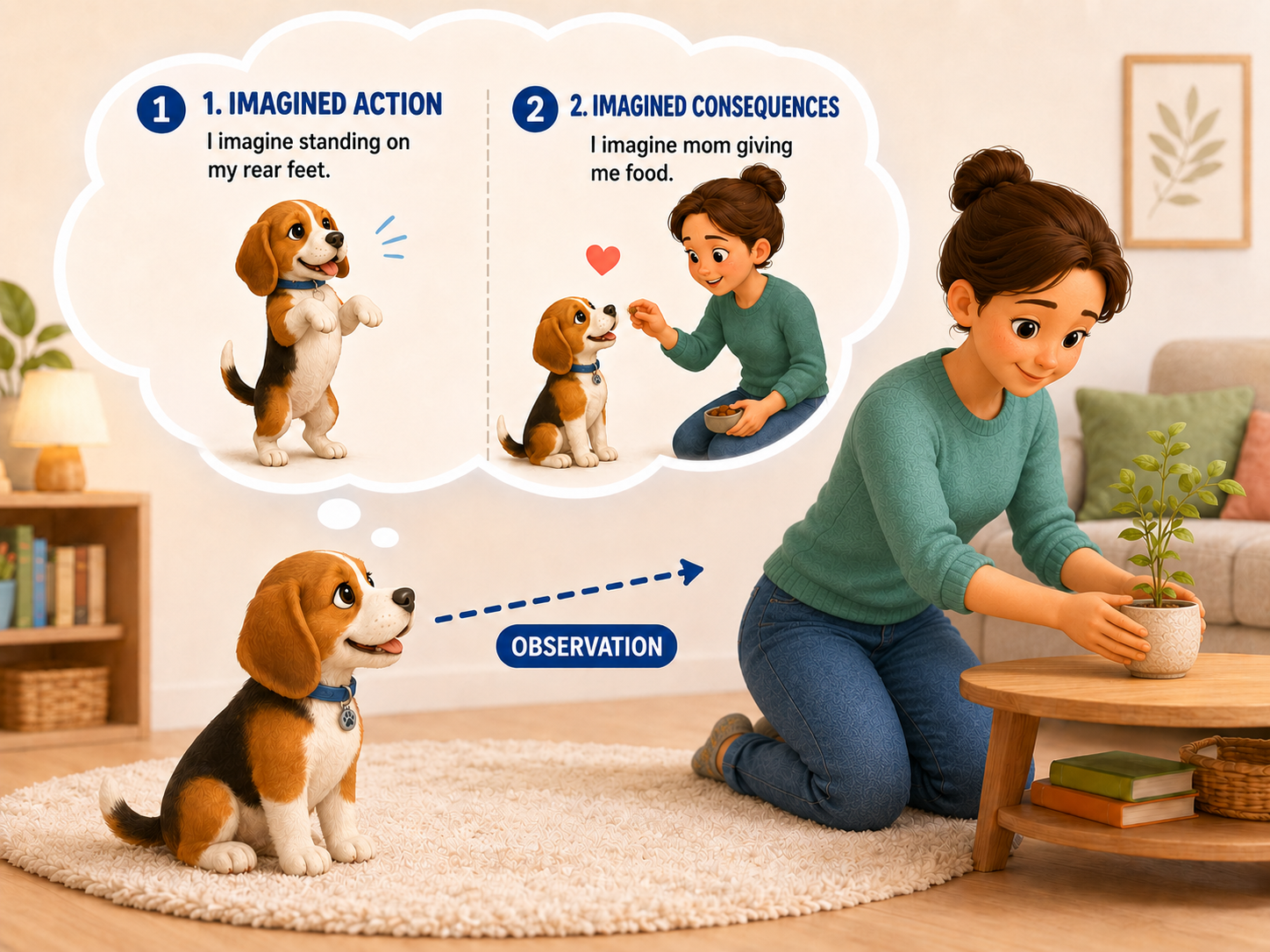}
    \end{subfigure}
    \caption{WMs are inspired by the implicit mental models humans and animals use to anticipate action consequences.}
    \label{fig:wm_inspire}
\end{figure}

World models draw inspiration from the internal mental representations that humans and animals use to interpret their surroundings and guide behavior \cite{ha2018recurrent,lecun2022path}. \Cref{fig:wm_inspire} illustrates this intuition through imagined actions and their possible consequences: a child anticipates being reprimanded after breaking a vase, while a dog anticipates receiving food after standing on its rear feet. Ha and Schmidhuber \cite{ha2018recurrent} develop a computational approach that combines compressed visual representations, a recurrent predictive model, and a controller that uses these representations to select actions. LeCun \cite{lecun2022path} further emphasizes the role of predictive world models in evaluating possible futures and planning toward objectives.

\begin{figure}[htpb!]
        \centering
        \includegraphics[width=0.62\linewidth]{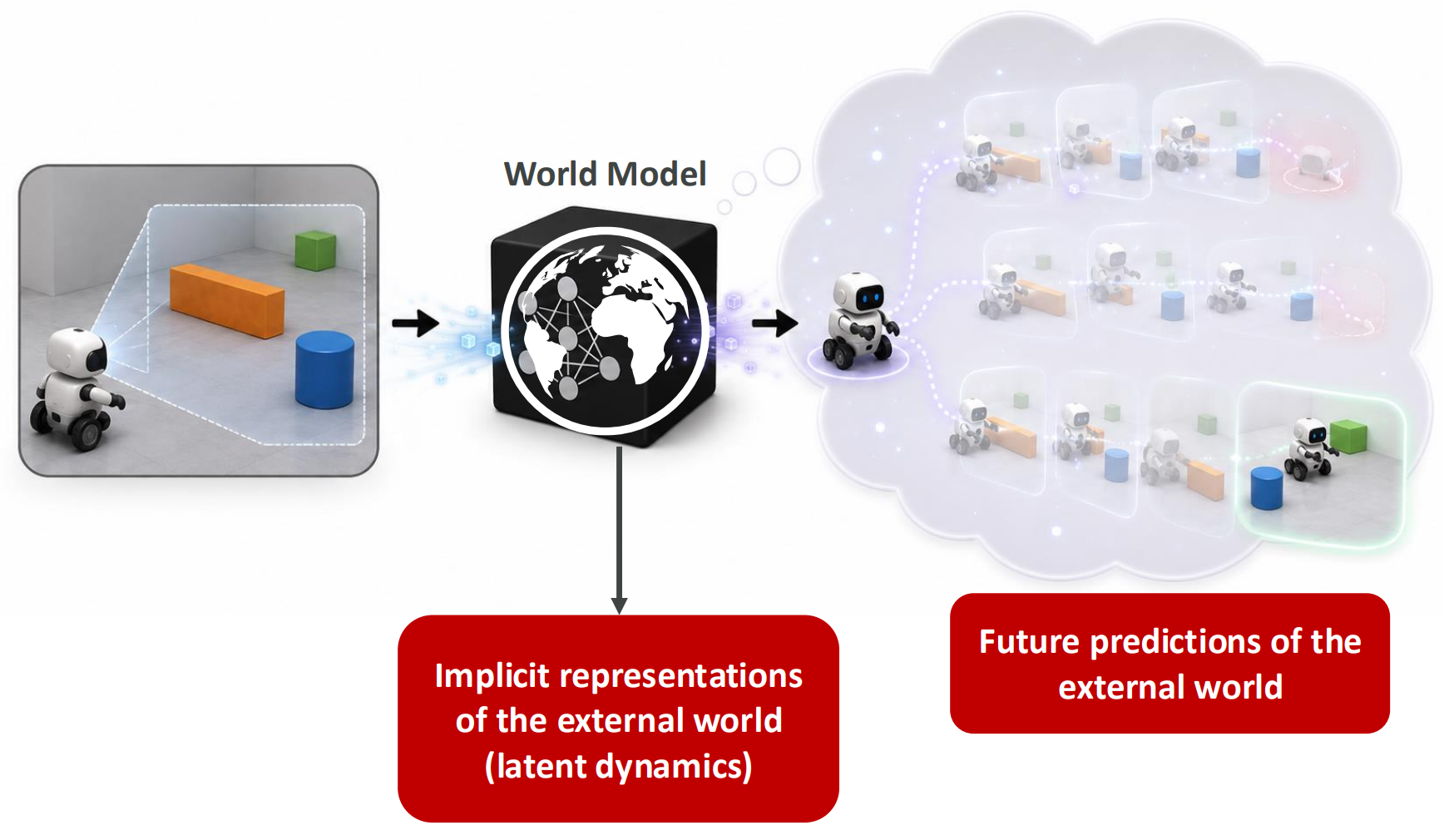}
       \caption{World models enable embodied systems to anticipate action consequences and plan ahead.}
        \label{fig:wm3}
\end{figure}

\Cref{fig:wm3} illustrates how a world model connects observations to prediction and decision-making. The robot forms an internal representation of its surroundings and predicts how they may evolve under candidate actions. These predictions allow it to evaluate possible outcomes and select actions toward its objective before executing them in the physical environment.

\subsection{Pixel-based and vector-based observations.}
\begin{figure}[htpb!]
    \centering
    \includegraphics[width=0.8\linewidth]{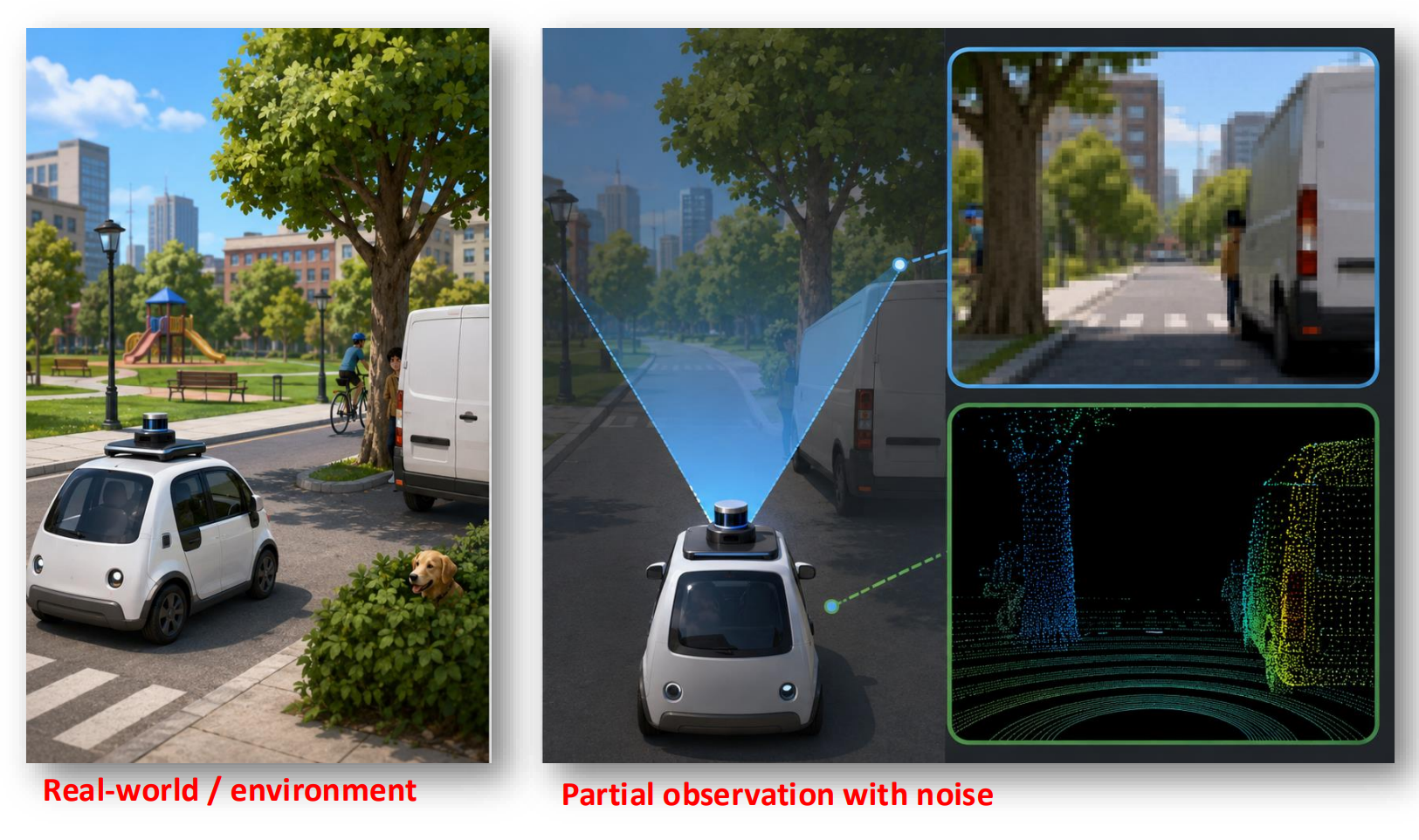}
    \caption{An illustration of pixel-based partial observation in a real-world setting.}
    \label{fig:po}
\end{figure}

Pixel-based observations represent visual information as images or video frames. For example, a robot's camera observation can be written as $y(t)=P(t)$, where $P(t)\in\mathbb{R}^{m\times n}$ is a grayscale image or $P(t)\in\mathbb{R}^{m\times n\times 3}$ is a color image at time $t$. Individual images generally provide only partial information about the underlying environment state \cite{hafner2019learning,lee2020stochastic}, as illustrated in \Cref{fig:po}.

Vector-based observations consist of numerical sensor readings arranged in a vector. For example, a wheeled robot's observation may be written as $y(t)=[p_x(t),p_y(t)]^\top\in\mathbb{R}^{2}$, containing its planar position. This constitutes partial observation of the robot's pose because its heading, $\theta(t)$, is not observed.

\section{Discussion and Future Work}\label{sub:diss}

K-PWM is motivated by the observation that many physical systems differ from
pixel-based world-model benchmarks: measurements are often vector-valued,
partially observed, and generated by sensors such as joint encoders, IMUs, or
force/torque sensors. In such settings, learning a useful world model requires
not only accurate prediction, but also inference over unobserved state variables
and calibrated uncertainty propagation over long horizons. By using the learned
state-space model itself for latent inference, K-PWM provides a direct connection
between prediction, filtering/smoothing, and downstream control.

Our empirical results indicate that Koopman-inspired structure remains valuable
even when combined with nonlinear neural decoders. The bilinear latent transition
offers a compact control-dependent dynamics model, while the decoder captures
nonlinear observation maps. This separation preserves analytical structure in the
latent dynamics without restricting the observation model to be linear. The
strong prediction performance in low-data regime, suggests
that such structure can improve both sample efficiency.

For control, embedding K-PWM inside MPPI provides a direct test of whether the
learned belief dynamics are useful beyond open-loop forecasting. The controller
uses the inferred belief state and propagates candidate action sequences through
the learned model to track task-relevant observation references. Although this is
not a full reward-maximizing reinforcement-learning setup, it demonstrates that
the learned partially observed world model can support closed-loop planning
across robotic tasks with different dynamics. This suggests a path toward
control-oriented world models that combine probabilistic inference, structured
latent dynamics, and sampling-based planning.

Several directions remain for future work. First, while the present method uses
Gaussian approximate inference through EKF/EKS, more expressive inference
schemes such as particle filtering, particle smoothing, unscented Kalman
filtering, or ensemble-based filtering may better capture multimodal or strongly
non-Gaussian posteriors. Second, extending K-PWM from reference-tracking MPPI to
reward-driven model-based reinforcement learning would allow direct comparison
with policy-learning and latent world-model baselines. Third, online adaptation
and uncertainty-aware exploration could further improve robustness when the test
environment differs from the training distribution. Finally, deploying \kpwm{} on
hardware systems with sensor noise, actuator delay, contact uncertainty, and
safety constraints would provide an important test of whether structured
probabilistic world models can transfer from simulation to real robotic systems.

\section{Limitations}\label{sub:limit}

Although K-PWM shows strong empirical performance, several limitations remain.
First, latent-state inference is based on a Gaussian approximation obtained using
an extended Kalman smoother. This approximation is computationally tractable and
well matched to the locally linearized decoder, but it may become inaccurate when
the posterior is highly non-Gaussian, multimodal, or strongly affected by
nonlinear observation ambiguities. Thus, the present formulation is best viewed
as an approximate inference method rather than an exact EM procedure.

Second, \kpwm{} does not constrain the stability of the learned latent transition. Stability can be important for long horizon predictions, which the GECM procedure doesn't ensure. Although, regularization helps.

Finally, 
real robotic systems introduce additional challenges, including unmodeled
actuator delays, sensor bias, domain shift, contact uncertainty, and safety
constraints during data collection. Extending K-PWM to online adaptation,
non-Gaussian filtering, richer uncertainty-aware planning objectives, and
hardware experiments is an important direction for future work.

\section{Ablations} \label{sub:ablation}

\paragraph{Effect of structured Koopman initialization.}
We evaluate whether the benefit of the structured Koopman initialization comes
only from selecting an appropriate latent hidden dimension, or also from
providing a useful initialization of the latent dynamics. For both Swimmer and
HalfCheetah, we keep the same hidden dimension as that selected by the structured
Koopman GECM initialization, but randomly initialize the generator matrices. All
other training settings are kept fixed. In both tasks, the randomly initialized
models reach the maximum GECM iteration budget without satisfying the convergence
criterion. By contrast, the structured Koopman initialization converged in lower GECM iterations.
These results indicate that the structured initialization is important not only
for model-size selection, but also for placing the Koopman-bilinear latent
dynamics in a favorable and well-conditioned region of the parameter space,
leading to more stable and substantially faster GECM convergence.

\paragraph{Smaller latent dimensions.}
We also evaluate whether K-PWM can be trained with smaller latent dimensions
than those selected by the structured Koopman initialization. For HalfCheetah, we train models with smaller latent dimensions using the same training-set size, $R=15$, and the
same optimization settings. In all cases, GECM reaches the maximum iteration
budget without satisfying the convergence criterion. By contrast, the latent
dimension selected by the structured Koopman initialization leads to successful
convergence. This ablation suggests that the proposed initialization is useful
not only for parameter initialization, but also for selecting a sufficiently
expressive latent dimension for stable EM-based learning under partial
observability.

These ablations evaluate convergence stability rather than final prediction
accuracy, since the randomly initialized and smaller-dimensional variants did
not satisfy the convergence criterion.

\paragraph{GECM versus joint gradient-based training.}
To assess the benefit of the GECM procedure in \Cref{alg:gecm_kpwm}, we train
the same \kpwm{} model by jointly optimizing all parameters
$\Theta=\{\widetilde\Theta,\bs\zeta\}$ with stochastic gradient descent (SGD),
instead of alternating the closed-form CM-step~1 updates with the
gradient-based decoder update. GECM attains prediction accuracy comparable to
joint SGD while training several times faster in wall-clock time. Moreover,
GECM avoids the learning-rate tuning and gradient clipping that joint SGD
requires for stable training, since the generators, noise covariances, and
latent prior are updated in closed form from the smoothed posterior moments.

\paragraph{Bilinear versus linear latent transition.}
We next isolate the role of the action-bilinear structure of the KBT in
\eqref{eq:HMM}. With every component other than the latent transition held
fixed, we replace the Koopman-bilinear transition by its linear submodel, in
which the action enters the latent dynamics additively rather than
multiplicatively. The bilinear transition consistently achieves substantially
lower prediction RMSE than its linear counterpart, indicating that the
multiplicative action dependence captures control effects that a
linear-in-the-action transition cannot.

\paragraph{Model-induced versus amortized inference.}
Finally, we evaluate what \kpwm{} gains from inferring latent states through
the learned model itself. We replace the EKF/EKS inference, which introduces
no additional trainable parameters, by a parameter-matched GRU recognition
network in the style of DVBF~\cite{karl2017deep} and deep Kalman filters,
holding the transition, decoder, latent dimension, training data, and rollout
protocol fixed. Replacing the inference mechanism changes the full-window
prediction RMSE only marginally, with no consistent winner across tasks and
data regimes. Thus, \kpwm{} matches the predictive performance of amortized
inference without a separately trained recognition network, without KL
warm-up, and without an additional inference training budget.

\newpage

\section{Supplementary Information for \Cref{sec:kpwm}} \label{sub:gecm}

\subsection{Statement and Proof of \Cref{prop:1}.}

\subsubsection{Complete statement of \Cref{prop:1}}\label{prop1:statement}

With \Aref{assum:ps} satisfied, at the \(j\)-th EM iteration, CM-step 1 yields \(\widetilde{\Theta}_j
=
\{\mf V_{0,j},\ldots,\mf V_{m,j},\bs\Sigma_{w,j},\bs\Sigma_{v,j},\bs\Sigma_{0,j},\bs\mu_{0,j}\}\) as the explicit maximizer of the ELBO, given by
\salnn{eq:exp_cm1}{
& \begin{bmatrix}
\mf V_{0,j}^{\top}
&
\mf V_{1,j}^{\top}
&
\cdots
&
\mf V_{m,j}^{\top}
\end{bmatrix}
=
\mf D \mf M^{-1}, \ \bs\mu_{0,j}
= \tfrac{1}{R}
{\scriptstyle \sum}_{r=1}^{R} \mu_0^{(r)},\\
& \bs\Sigma_{0,j}
=
\tfrac{1}{R}
{\scriptstyle \sum}_{r=1}^{R}
\left[
\bs\Sigma_{0,0}^{(r)}
+
\left(
\bs\mu_0^{(r)}
-
\bs\mu_{0,j}
\right)
\left(
\bs\mu_0^{(r)}
-
\bs\mu_{0,j}
\right)^\top
\right] \\
& \bs\Sigma_{w,j}
=
\tfrac{1}{RK}
{\scriptstyle \sum}_{r=1}^{R}
{\scriptstyle \sum}_{k=0}^{K-1}
\Big[
\bs\Sigma_{k+1,k+1}^{(r)}
-
\mf A_{k,j}^{(r)}\bs\Sigma_{k,k+1}^{(r)}
-
\bs\Sigma_{k+1,k}^{(r)}\mf A_{k,j}^{(r)\top}
+
\mf A_{k,j}^{(r)}
\bs\Sigma_{k,k}^{(r)}
\mf A_{k,j}^{(r)\top}
+ \notag\\
& \qquad \qquad \qquad \qquad \qquad \left(
\mu_{k+1}^{(r)}
-
\mf A_{k,j}^{(r)}\mu_k^{(r)}
-
\bs b_{k,j}^{(r)}
\right)
\left(
\mu_{k+1}^{(r)}
-
\mf A_{k,j}^{(r)}\mu_k^{(r)}
-
\bs b_{k,j}^{(r)}
\right)^\top
\Big],\\
& \bs\Sigma_{v,j}
=
\tfrac{1}{R(K+1)}
{\scriptstyle \sum}_{r=1}^{R}
{\scriptstyle \sum}_{k=0}^{K}
\Big[
\mf J_{k,j}^{(r)}
\bs\Sigma_{k,k}^{(r)}
\mf J_{k,j}^{(r)\top}
+
\big( y_k^{(r)} - h_{\bs \zeta_{j-1}}(\mu_k^{(r)})\big)
\big( y_k^{(r)} - h_{\bs \zeta_{j-1}}(\mu_k^{(r)})\big)^\top
\Big],
}
where \(h_{\bs\zeta_{j-1}}\) explicitly represents the decoder whose weights and biases are contained in \(\bs\zeta_{j-1}\) and \(\mf J_{k,j}^{(r)} := \nabla_{ z} h_{\bs\zeta_{j-1}}( z)\big|_{ z=\mu_k^{(r)}}\).

\subsubsection{Proof of \Cref{prop:1}}\label{proof:prop1} 

Since the entropy term
\(-\sum_{r=1}^R \E_{q_j^{(r)}}[\log q_j^{(r)}(Z^{(r)})]\) in
\eqref{eq:ELBO1} does not depend on \(\widetilde \Theta\), CM-step 1 only
requires maximizing
\[
{\scriptstyle \sum}_{r=1}^{R}
\E_{q_j^{(r)}}\!\left[
\log p_{\Theta}(Z^{(r)},Y^{(r)}\mid \mcr A^{(r)})
\right]
\]
with respect to
\(\widetilde \Theta=
\{\mf V_0,\ldots,\mf V_m,\bs\Sigma_w,\bs\Sigma_v,\bs\Sigma_0,\mu_0\}\).
Using the factorization in \eqref{eq:joint_density}, the terms depending on
\(\widetilde \Theta\) split into the initial-prior term, the transition term,
and the observation term. Up to constants independent of
\(\widetilde \Theta\), this objective is
\begin{align}
\mcr Q(\widetilde \Theta)
&=
-\frac{R}{2}\log\det \bs\Sigma_0
-\frac{1}{2}
{\scriptstyle \sum}_{r=1}^{R}
\E_{q_j^{(r)}}\!\left[
\|z_0^{(r)}-\mu_0\|_{\bs\Sigma_0^{-1}}^2
\right]
\notag\\
&\quad
-\frac{RK}{2}\log\det \bs\Sigma_w
-\frac{1}{2}
{\scriptstyle \sum}_{r=1}^{R}
{\scriptstyle \sum}_{k=0}^{K-1}
\E_{q_j^{(r)}}\!\left[
\|z_{k+1}^{(r)}
-\mf A_k^{(r)}z_k^{(r)}
-\bs b_k^{(r)}\|_{\bs\Sigma_w^{-1}}^2
\right]
\notag\\
&\quad
-\frac{R(K+1)}{2}\log\det \bs\Sigma_v
-\frac{1}{2}
{\scriptstyle \sum}_{r=1}^{R}
{\scriptstyle \sum}_{k=0}^{K}
\E_{q_j^{(r)}}\!\left[
\|y_k^{(r)}-h_{\bs\zeta_{j-1}}(z_k^{(r)})\|_{\bs\Sigma_v^{-1}}^2
\right].
\label{eq:proof_q_function}
\end{align}
Here \(\mf A_k^{(r)}\) and \(\bs b_k^{(r)}\) are determined by
\[
\bmat{1 & \bs b_k^{(r)\top}\\ \mf 0 & \mf A_k^{(r)\top}}
=
\mf I+\Delta t
{\scriptstyle \sum}_{i=0}^{m}
a_{k_i}^{(r)}\mfk G_i,
\qquad
\mfk G_i=\bmat{\mf 0 & \mf V_i}.
\]

We first maximize with respect to the generator matrices. Let
\[
\mf V^\top :=
\bmat{
\mf V_0^\top & \mf V_1^\top & \cdots & \mf V_m^\top
},
\qquad
\chi_k^{(r)} :=
\bs a_k^{(r)}\kron \mf z_k^{(r)},
\qquad
\dot z_k^{(r)}
:=
\frac{z_{k+1}^{(r)}-z_k^{(r)}}{\Delta t}.
\]
Then the Koopman-bilinear transition can be written in differential form as
\[
\dot z_k^{(r)}
=
\mf V^\top \chi_k^{(r)}
+
\text{noise}.
\]
For fixed \(\bs\Sigma_w\), maximizing the transition contribution in
\eqref{eq:proof_q_function} with respect to \(\mf V\) is equivalent to
minimizing
\[
\Phi(\mf V)
=
{\scriptstyle \sum}_{r=1}^{R}
{\scriptstyle \sum}_{k=0}^{K-1}
\E_{q_j^{(r)}}\!\left[
\left\|
\dot z_k^{(r)}
-
\mf V^\top \chi_k^{(r)}
\right\|_{\bs\Sigma_w^{-1}}^2
\right].
\]
Using the definitions of \(\mf M\) and \(\mf D\), this objective can be written as
\[
\Phi(\mf V)
=
\operatorname{const}
-
2\trace\!\left(
\bs\Sigma_w^{-1}\mf D\mf V
\right)
+
\trace\!\left(
\bs\Sigma_w^{-1}\mf V^\top \mf M \mf V
\right).
\]
Taking a first variation in the direction \(\delta \mf V\) gives
\[
\delta \Phi(\mf V)
=
2\trace\!\left[
\bs\Sigma_w^{-1}
\left(
\mf V^\top \mf M-\mf D
\right)
\delta \mf V
\right].
\]
Since \(\bs\Sigma_w\succ 0\), the stationary condition is
\[
\mf V^\top \mf M=\mf D.
\]
By \Aref{assum:ps}, \(\mf M\succ 0\), hence \(\mf M\) is invertible and the
unique stationary point is
\[
\mf V^\top
=
\mf D\mf M^{-1}.
\]
Because \(\Phi(\mf V)\) is a strictly convex quadratic function when
\(\mf M\succ 0\), this stationary point is the unique minimizer of
\(\Phi\), and therefore the unique maximizer of the transition ELBO with
respect to the generator matrices. This proves the first update in
\eqref{eq:exp_cm1}.

Next, for fixed generator matrices, define the transition residual
\[
e_{k}^{(r)}
:=
z_{k+1}^{(r)}
-
\mf A_{k,j}^{(r)}z_k^{(r)}
-
\bs b_{k,j}^{(r)} .
\]
The \(\bs\Sigma_w\)-dependent part of \(\mcr Q\) is
\[
-\frac{RK}{2}\log\det\bs\Sigma_w
-
\frac{1}{2}
\trace\!\left(
\bs\Sigma_w^{-1}
{\scriptstyle \sum}_{r=1}^{R}
{\scriptstyle \sum}_{k=0}^{K-1}
\E_{q_j^{(r)}}[e_k^{(r)}e_k^{(r)\top}]
\right).
\]
For a positive-definite covariance matrix, this Gaussian covariance
maximization has the unique solution
\[
\bs\Sigma_{w,j}
=
\frac{1}{RK}
{\scriptstyle \sum}_{r=1}^{R}
{\scriptstyle \sum}_{k=0}^{K-1}
\E_{q_j^{(r)}}[e_k^{(r)}e_k^{(r)\top}].
\]
Expanding the expectation using the smoothed first and second moments gives
\[
\begin{aligned}
\E_{q_j^{(r)}}[e_k^{(r)}e_k^{(r)\top}]
&=
\bs\Sigma_{k+1,k+1}^{(r)}
-
\mf A_{k,j}^{(r)}\bs\Sigma_{k,k+1}^{(r)}
-
\bs\Sigma_{k+1,k}^{(r)}\mf A_{k,j}^{(r)\top}
+
\mf A_{k,j}^{(r)}
\bs\Sigma_{k,k}^{(r)}
\mf A_{k,j}^{(r)\top}
\\
&\quad
+
\left(
\mu_{k+1}^{(r)}
-
\mf A_{k,j}^{(r)}\mu_k^{(r)}
-
\bs b_{k,j}^{(r)}
\right)
\left(
\mu_{k+1}^{(r)}
-
\mf A_{k,j}^{(r)}\mu_k^{(r)}
-
\bs b_{k,j}^{(r)}
\right)^\top .
\end{aligned}
\]
Substituting this expression yields the stated update for
\(\bs\Sigma_{w,j}\).

We now maximize with respect to the initial prior parameters. The prior
contribution is
\[
-\frac{R}{2}\log\det\bs\Sigma_0
-
\frac{1}{2}
{\scriptstyle \sum}_{r=1}^{R}
\E_{q_j^{(r)}}\!\left[
(z_0^{(r)}-\mu_0)^\top
\bs\Sigma_0^{-1}
(z_0^{(r)}-\mu_0)
\right].
\]
For fixed \(\bs\Sigma_0\), differentiating with respect to \(\mu_0\) gives
\[
{\scriptstyle \sum}_{r=1}^{R}
\bs\Sigma_0^{-1}
\left(
\mu_0^{(r)}-\mu_0
\right)
=
0,
\]
and hence
\[
\bs\mu_{0,j}
=
\frac{1}{R}
{\scriptstyle \sum}_{r=1}^{R}
\mu_0^{(r)}.
\]
For fixed \(\bs\mu_{0,j}\), the Gaussian covariance maximizer is
\[
\bs\Sigma_{0,j}
=
\frac{1}{R}
{\scriptstyle \sum}_{r=1}^{R}
\E_{q_j^{(r)}}\!\left[
(z_0^{(r)}-\bs\mu_{0,j})
(z_0^{(r)}-\bs\mu_{0,j})^\top
\right].
\]
Expanding this expectation gives
\[
\bs\Sigma_{0,j}
=
\tfrac{1}{R}
{\scriptstyle \sum}_{r=1}^{R}
\left[
\bs\Sigma_{0,0}^{(r)}
+
\left(
\bs\mu_0^{(r)}
-
\bs\mu_{0,j}
\right)
\left(
\bs\mu_0^{(r)}
-
\bs\mu_{0,j}
\right)^\top
\right],
\]
which is the stated update.

Finally, we maximize with respect to the observation covariance
\(\bs\Sigma_v\). Since the decoder is nonlinear, we evaluate the observation
term under the same first-order Gaussian approximation used in the EKS. Namely,
around the smoothed mean \(\mu_k^{(r)}\),
\[
h_{\bs\zeta_{j-1}}(z_k^{(r)})
\approx
h_{\bs\zeta_{j-1}}(\mu_k^{(r)})
+
\mf J_{k,j}^{(r)}
\left(
z_k^{(r)}-\mu_k^{(r)}
\right),
\qquad
\mf J_{k,j}^{(r)}
=
\nabla_z h_{\bs\zeta_{j-1}}(z)\big|_{z=\mu_k^{(r)}} .
\]
Therefore,
\[
\begin{aligned}
&\E_{q_j^{(r)}}\!\left[
\left(
y_k^{(r)}-h_{\bs\zeta_{j-1}}(z_k^{(r)})
\right)
\left(
y_k^{(r)}-h_{\bs\zeta_{j-1}}(z_k^{(r)})
\right)^\top
\right]
\\
&\qquad \approx
\mf J_{k,j}^{(r)}
\bs\Sigma_{k,k}^{(r)}
\mf J_{k,j}^{(r)\top}
+
\left(
y_k^{(r)}
-
h_{\bs\zeta_{j-1}}(\mu_k^{(r)})
\right)
\left(
y_k^{(r)}
-
h_{\bs\zeta_{j-1}}(\mu_k^{(r)})
\right)^\top .
\end{aligned}
\]
The \(\bs\Sigma_v\)-dependent part of the approximate observation ELBO is
\[
-\frac{R(K+1)}{2}\log\det\bs\Sigma_v
-
\frac{1}{2}
\trace\!\left(
\bs\Sigma_v^{-1}\mf S_v
\right),
\]
where
\[
\mf S_v
=
{\scriptstyle \sum}_{r=1}^{R}
{\scriptstyle \sum}_{k=0}^{K}
\left[
\mf J_{k,j}^{(r)}
\bs\Sigma_{k,k}^{(r)}
\mf J_{k,j}^{(r)\top}
+
\left(
y_k^{(r)}
-
h_{\bs\zeta_{j-1}}(\mu_k^{(r)})
\right)
\left(
y_k^{(r)}
-
h_{\bs\zeta_{j-1}}(\mu_k^{(r)})
\right)^\top
\right].
\]
The unique positive-definite maximizer is therefore
\[
\bs\Sigma_{v,j}
=
\frac{1}{R(K+1)}\mf S_v,
\]
which is exactly the stated update.

Under \Aref{assum:ps}, the matrices \(\mf M\), \(\bs\Sigma_w\),
\(\bs\Sigma_v\), and \(\bs\Sigma_0\) are positive definite. Hence the quadratic
generator objective is strictly convex, and the Gaussian covariance objectives
have unique positive-definite maximizers. Therefore, the updates in
\eqref{eq:exp_cm1} give the explicit CM-step 1 maximizer of the ELBO with
respect to \(\widetilde \Theta\), with the observation term interpreted under the
EKS linearized Gaussian approximation.

\subsection{Supplementary info on \Aref{assum:ps}}\label{sub:assumpe}

\paragraph{Screening for persistency of excitation (PE).}
Consider the action sequence $a^{(r)}_{0:K-1}$ of the $r$-th candidate
training trajectory. Let $\bar a^{(r)}_i$ and $s^{(r)}_i$ denote the
sample mean and standard deviation of action component $i\in[1:m]$,
respectively. Candidates with any $s^{(r)}_i=0$ are rejected before
standardization. For the remaining candidates, we define
$\tilde a^{(r)}_{k,i}=(a^{(r)}_{k,i}-\bar a^{(r)}_i)/s^{(r)}_i$
and apply the criterion
\begin{equation}
  \lambda_{\min}\!\big(\boldsymbol{\Gamma}^{(r)}\big) \ge \lambda_{\mathrm{PE}},
  \qquad
  \boldsymbol{\Gamma}^{(r)} := \tfrac{1}{K}{\scriptstyle \sum}_{k=0}^{K-1}\tilde a^{(r)}_k\,\tilde a^{(r)\top}_k ,
  \label{eq:pe_test}
\end{equation}
where $\lambda_{\mathrm{PE}}>0$ is a fixed threshold. The test is
invariant to nonzero componentwise rescaling of the actions and
requires the mean squared projection of the standardized actions onto
every unit direction to be at least $\lambda_{\mathrm{PE}}$. This bound
is sufficient for order-one PE of the standardized sequence in the
sense of \citet{willems2005note}. Whenever a candidate is rejected,
we proceed to the next seed.

\paragraph{Ensuring positive definiteness of covariances.}
At initialization and after each CM-step~1, we apply eigenvalue
floors $\gamma_v>0$, $\gamma_w>0$, and $\gamma_0>0$ to the
observation-noise, process-noise, and initial-state covariance
estimates, respectively. For each $i\in\{v,w,0\}$, let
$U_i\operatorname{diag}(\lambda_{i,1},\ldots,\lambda_{i,d_i})U_i^\top$
be the eigendecomposition of the symmetrized covariance estimate.
We set
\[
\Sigma_i
=
U_i\operatorname{diag}\!\left(
\max\{\lambda_{i,1},\gamma_i\},\ldots,
\max\{\lambda_{i,d_i},\gamma_i\}
\right)U_i^\top.
\]
This replaces eigenvalues below $\gamma_i$ with $\gamma_i$ and
ensures $\Sigma_i\succeq\gamma_i I\succ0$ throughout training.
In exact arithmetic, these positive-definite model covariances yield
positive-definite smoothed latent covariances in the EKF--RTS E-step.
The corresponding augmented latent second moments are positive
definite by the Schur complement. 

\paragraph{Generator regularization and conditioning of \(\mf M\).}
In CM-step~1 the generator blocks are obtained from the ridge-regularized normal equations
\[
\begin{bmatrix}
\mf V_{0,j}^{\top} & \mf V_{1,j}^{\top} & \cdots & \mf V_{m,j}^{\top}
\end{bmatrix}
=
\mf D\,\big(\mf M + \gamma_G \mf I\big)^{-1},
\qquad \gamma_G > 0,
\]
which maximizes the regularized ELBO $\widehat{\mcr F}$ with the penalty
$\tfrac{\gamma_G}{2}\,{\scriptstyle\sum}_{i=0}^{m}\|\bs\Sigma_w^{-1/2}\mf V_i^{\top}\|_F^2$ on the
generators; the corresponding process-noise update acquires the additional term
$\tfrac{\gamma_G}{RK}\,\mf W\mf W^\top$, with
$\mf W = [\mf V_0^\top \ \cdots \ \mf V_m^\top]$. We use $\gamma_G = 10^{-3}$, except
$10^{-2}$ for HalfCheetah, $10^{-4}$ for TIAGo++ and $10^{-6}$ for Allegro Hand.
Since $\mf M \succeq 0$ (a sum of Kronecker products of positive semidefinite
matrices), the ridge alone guarantees $\mf M + \gamma_G\mf I \succeq \gamma_G \mf I$,
and therefore a well-defined update with condition number at most
$\big(\lambda_{\max}(\mf M)+\gamma_G\big)/\big(\lambda_{\min}(\mf M)+\gamma_G\big)$.
Together with
the PE screen and the eigenvalue floors that enforce \Aref{assum:ps}, $\mf M$ is in fact
positive definite, so that the ridge acts only as a safeguard against round-off and
near-collinear latent features rather than as a substitute for missing excitation:
\begin{enumerate}
\item \emph{Smoothed covariances are positive definite.} With $\bs\Sigma_w \succeq \gamma_w\mf I$,
$\bs\Sigma_v \succeq \gamma_v\mf I$ and $\bs\Sigma_0 \succeq \gamma_0\mf I$, the EKF gives
$\bs\Sigma_{k+1|k} = \mf A_k\bs\Sigma_{k|k}\mf A_k^\top + \bs\Sigma_w \succeq \gamma_w\mf I$ and
$\bs\Sigma_{k+1|k+1} = \big(\bs\Sigma_{k+1|k}^{-1} + \mf J_{k+1}^\top\bs\Sigma_v^{-1}\mf J_{k+1}\big)^{-1} \succ 0$,
starting from $\bs\Sigma_0 \succ 0$. The Rauch--Tung--Striebel recursion can be written as
$\bs\Sigma_{k,k} = \big(\bs\Sigma_{k|k}^{-1} + \mf A_k^\top\bs\Sigma_w^{-1}\mf A_k\big)^{-1}
+ \mf C_k\bs\Sigma_{k+1,k+1}\mf C_k^\top$, a positive definite matrix plus a positive
semidefinite one; hence $\bs\Sigma_{k,k}^{(r)} \succ 0$ for all $k$ and $r$.
\item \emph{Augmented second moments are positive definite.} Since
\[
\mc M_k^{(r)}
=
\begin{bmatrix}
1 & \mu_k^{(r)\top} \\
\mu_k^{(r)} & \mu_k^{(r)}\mu_k^{(r)\top} + \bs\Sigma_{k,k}^{(r)}
\end{bmatrix},
\]
whose Schur complement with respect to the leading entry is $\bs\Sigma_{k,k}^{(r)} \succ 0$, we have
$\mc M_k^{(r)} \succ 0$. Over the finitely many pairs $(r,k)$,
$\beta := \min_{r,k}\lambda_{\min}\big(\mc M_k^{(r)}\big) > 0$.
\item \emph{PE gives full rank of the action Gram matrix.} The PE screen \eqref{eq:pe_test} ensures
that the standardized actions of every training trajectory have a positive definite sample
covariance. By the Schur complement with respect to the constant entry, this is equivalent to
$\mf G_a := {\scriptstyle\sum}_{r=1}^{R}{\scriptstyle\sum}_{k=0}^{K-1}\bs a_k^{(r)}\bs a_k^{(r)\top} \succ 0$
for the augmented actions $\bs a_k^{(r)} = [1 \ \ a_k^{(r)\top}]^\top$.
\item \emph{Conclusion.} Since $\bs a_k^{(r)}\bs a_k^{(r)\top} \succeq 0$ and
$\mc M_k^{(r)} - \beta\mf I \succeq 0$, and Kronecker products of positive semidefinite matrices are
positive semidefinite,
\[
\mf M
=
{\scriptstyle\sum}_{r,k}\big(\bs a_k^{(r)}\bs a_k^{(r)\top}\big)\otimes\mc M_k^{(r)}
\;\succeq\;
\beta\,\mf G_a \otimes \mf I
\;\succeq\;
\beta\,\lambda_{\min}(\mf G_a)\,\mf I
\;\succ\; 0 .
\]
\end{enumerate}
Hence, in exact arithmetic, \Aref{assum:ps} (enforced by the eigenvalue floors) and the PE screen
together guarantee $\mf M \succ 0$, so the unregularized maximizer $\mf D\mf M^{-1}$ of
\Cref{prop:1} exists; the generator ridge additionally bounds the condition number of the
system that is solved in practice.

\subsection{Proof of \Pref{prop:2}}\label{proof:prop2}
Fix \(Q_j=\{q_j^{(r)}\}_{r=1}^R\) and
\(\widetilde{\Theta}_j\). In particular, during CM-step 2 the transition
parameters, the initial-state distribution, and the covariance matrices
\(\bs\Sigma_{w,j},\bs\Sigma_{v,j},\bs\Sigma_{0,j}\) are fixed. Hence the only
part of the ELBO that depends on the decoder parameter \(\bs\zeta\) is the
expected observation log-likelihood.

Using the complete-data factorization in \eqref{eq:joint_density}, the ELBO can
be written as
\begin{align}
\mcr F\big((\widetilde{\Theta}_j,\bs\zeta),Q_j\big)
&=
{\scriptstyle \sum}_{r=1}^{R}
\E_{q_j^{(r)}}\!\left[
\begin{aligned}
&
\log p_{\widetilde{\Theta}_j}(z_0^{(r)})
+
{\scriptstyle \sum}_{k=0}^{K-1}
\log p_{\widetilde{\Theta}_j}
\!\left(
z_{k+1}^{(r)}
\mid
z_k^{(r)},a_k^{(r)}
\right)
\\
& \qquad +
{\scriptstyle \sum}_{k=0}^{K}
\log p_{\widetilde{\Theta}_j,\bs\zeta}
\!\left(
y_k^{(r)}
\mid
z_k^{(r)}
\right)
-
\log q_j^{(r)}(Z^{(r)})
\end{aligned}
\right].
\end{align}
Since \(Q_j\) and \(\widetilde{\Theta}_j\) are fixed, the prior term, transition
term, and entropy term are independent of \(\bs\zeta\). Therefore, there exists
a constant \(C_j\), independent of \(\bs\zeta\), such that
\[
\mcr F\big((\widetilde{\Theta}_j,\bs\zeta),Q_j\big)
=
C_j
+
{\scriptstyle \sum}_{r=1}^{R}
{\scriptstyle \sum}_{k=0}^{K}
\E_{q_j^{(r)}}\!\left[
\log p_{\widetilde{\Theta}_j,\bs\zeta}
\!\left(
y_k^{(r)}
\mid
z_k^{(r)}
\right)
\right].
\]

Under the Gaussian observation model in \eqref{eq:rec_model},
\[
p_{\widetilde{\Theta}_j,\bs\zeta}
\!\left(
y_k^{(r)}
\mid
z_k^{(r)}
\right)
\approx
\Nc\!\left(
y_k^{(r)};
h_{\bs\zeta}(z_k^{(r)}),
\bs\Sigma_{v,j}
\right).
\]
Thus,
\[
\log p_{\widetilde{\Theta}_j,\bs\zeta}
\!\left(
y_k^{(r)}
\mid
z_k^{(r)}
\right)
=
-\frac{1}{2}
\big\|
y_k^{(r)}
-
h_{\bs\zeta}(z_k^{(r)})
\big\|_{\bs\Sigma_{v,j}^{-1}}^2
-\frac{1}{2}\log\det \bs\Sigma_{v,j}
-\frac{n_y}{2}\log(2\pi).
\]
Because \(\bs\Sigma_{v,j}\) is fixed in CM-step 2, the log-determinant and
normalization terms are also independent of \(\bs\zeta\). Hence, there exists
another constant \(\widetilde C_j\), independent of \(\bs\zeta\), such that
\[
\mcr F\big((\widetilde{\Theta}_j,\bs\zeta),Q_j\big)
=
\widetilde C_j
-
\frac{1}{2}
{\scriptstyle \sum}_{r=1}^{R}
{\scriptstyle \sum}_{k=0}^{K}
\E_{q_j^{(r)}}
\left[
\big\|
y_k^{(r)}
-
h_{\bs\zeta}(z_k^{(r)})
\big\|_{\bs\Sigma_{v,j}^{-1}}^2
\right].
\]
Using the definition of \(\mc L_{\mathrm{dec}}\), this becomes
\[
\mcr F\big((\widetilde{\Theta}_j,\bs\zeta),Q_j\big)
=
\widetilde C_j
-
\frac{R(K+1)}{2}
\mc L_{\mathrm{dec}}(\bs\zeta).
\]
Therefore,
\[
\begin{aligned}
&
\mcr F\big((\widetilde{\Theta}_j,\bs\zeta_j),Q_j\big)
-
\mcr F\big((\widetilde{\Theta}_j,\bs\zeta_{j-1}),Q_j\big)
\\
&\qquad =
-\frac{R(K+1)}{2}
\left[
\mc L_{\mathrm{dec}}(\bs\zeta_j)
-
\mc L_{\mathrm{dec}}(\bs\zeta_{j-1})
\right].
\end{aligned}
\]
If
\[
\mc L_{\mathrm{dec}}(\bs\zeta_j)
\le
\mc L_{\mathrm{dec}}(\bs\zeta_{j-1}),
\]
then
\[
-\frac{R(K+1)}{2}
\left[
\mc L_{\mathrm{dec}}(\bs\zeta_j)
-
\mc L_{\mathrm{dec}}(\bs\zeta_{j-1})
\right]
\ge 0.
\]
Consequently,
\[
\mcr F\big((\widetilde{\Theta}_j,\bs\zeta_j),Q_j\big)
\ge
\mcr F\big((\widetilde{\Theta}_j,\bs\zeta_{j-1}),Q_j\big).
\]
Thus, any decoder update that does not increase
\(\mc L_{\mathrm{dec}}\) performs a generalized conditional maximization step
with respect to the decoder block. This proves the claim.

\begin{remark}[Posterior-mean implementation of CM-step 2]
In implementation, CM-step 2 uses the smoothed posterior means
$\mu_{k|K}^{(r)}$ obtained from the EKF/Rauch smoother as the latent inputs to
the MLP decoder. This is a standard tractable choice for nonlinear observation
maps, where the exact expectation over the smoothing distribution is not
available in closed form. Thus, the decoder block is optimized as a supervised
regression problem from smoothed latent representations to observations, while
the smoothing covariances are still used in the observation-noise update.
\end{remark}

\begin{algorithm}[htpb!]
\caption{GECM Training of \kpwm{}}
\label{alg:gecm_kpwm}

\KwIn{Dataset $\mathcal D=\{(y^{(r)}_{0:K},a^{(r)}_{0:K-1})\}_{r=1}^R$,
maximum and minimum iterations $j_{\mrm{max}}$, $j_{\mrm{min}}$, tolerance $\epsilon$, patience $P$}
\KwOut{Learned parameters $\Theta^\star=\{\widetilde{\Theta}^\star,\bs\zeta^\star\}$}

Initialize $\Theta_0=\{\widetilde{\Theta}_0,\bs\zeta_0\}$ using Koopman-consistent initialization\;
Set $\ell^\star=-\infty$, $\Theta^\star=\Theta_0$, $p=0$\;

\For{$j=1,\ldots,j_{\mrm{max}}$}{

    \textbf{E-step: Latent-state inference}\;

    \For{$r=1,\ldots,R$}{
        Run EKF forward filtering using $\Theta_{j-1}$\;

        Run backward smoothing to obtain
        $q_j^{(r)}(z^{(r)}_{0:K})$\;

        Store the smoothing moments
        $\mu_k^{(r)}$, $\bs\Sigma_{k,k}^{(r)}$, and
        $\bs\Sigma_{k,k+1}^{(r)}$\;
    }

    \textbf{CM-step 1: Koopman-bilinear parameter update}\;

    Update the Koopman generator blocks using the smoothing moments:
    $
    \begin{bmatrix}
    \mf V_{0,j}^{\top} & \mf V_{1,j}^{\top} & \cdots & \mf V_{m,j}^{\top}
    \end{bmatrix}
    =
    \mf D \big(\mf M+\gamma_G\mf I\big)^{-1}
    $\;

    Update the latent prior parameters $\mu_{0,j}$ and $\bs\Sigma_{0,j}$\;

    Update the process-noise covariance $\bs\Sigma_{w,j}$ and observation-noise covariance $\bs\Sigma_{v,j}$
    (eigenvalue floors, \Aref{assum:ps})\;

    Set
    $
    \widetilde{\Theta}_j
    =
    \{\mf V_{0,j},\ldots,\mf V_{m,j},
    \bs\Sigma_{w,j},\bs\Sigma_{v,j},\bs\Sigma_{0,j},\mu_{0,j}\}.
    $\;

    \textbf{CM-step 2: Decoder update}\;

    With $\widetilde{\Theta}_j$ fixed, starting from $\bs\zeta_{j-1}$, train the MLP decoder using
    PyTorch backpropagation with a variable learning-rate schedule to obtain $\bs\zeta_j$\;

    Set $\Theta_j=\{\widetilde{\Theta}_j,\bs\zeta_j\}$\;

    Compute the regularized training log-likelihood per scalar
    $\ell_j=(\mcr L-\mcr R)(\Theta_j)/\big(R(K+1)n_y\big)$\;

    \eIf{$\ell_j>\ell^\star+\epsilon$}{
        Set $\ell^\star=\ell_j$, $\Theta^\star=\Theta_j$, $p=0$\;
    }{
        Set $p=p+1$\;
    }

    \If{$j\geq j_{\mrm{min}}$ \textbf{and} $p\geq P$}{
        \textbf{break}\;
    }
}

\Return{$\Theta^\star$}\;

\end{algorithm}

\subsection{Why EKS in the E-step?}

In the E-step of EM iteration \(j\), the parameters \(\Theta_{j-1}\) are fixed
and the auxiliary distribution \(Q\) is chosen to maximize the ELBO with respect
to \(Q\). By the standard EM identity,
\[
\mcr F(\Theta_{j-1},Q)
=
\log p_{\Theta_{j-1}}(Y\mid \mcr A)
-
D_{\mathrm{KL}}
\left(
Q(Z)
\,\middle\|\,
p_{\Theta_{j-1}}(Z\mid Y,\mcr A)
\right),
\]
the maximizing choice is
\[
Q_j(Z)
=
p_{\Theta_{j-1}}(Z\mid Y,\mcr A).
\]
Thus, the E-step amounts to latent-state smoothing under the current K-PWM
parameters.

For each trajectory \(r\), fixing \(\Theta_{j-1}\) and the action sequence
\(\mcr A^{(r)}\) gives a linear-Gaussian latent transition model. However, the
observation model remains nonlinear through the decoder
\[
y_k^{(r)}
=
h_{\bs\zeta_{j-1}}(z_k^{(r)})
+
v_k^{(r)}.
\]
As a result, the smoothing posterior
\[
p_{\Theta_{j-1}}(Z^{(r)}\mid Y^{(r)},\mcr A^{(r)})
\]
does not generally admit a closed-form Gaussian recursion. We therefore compute
the E-step using the Gaussian smoothing distribution induced by the extended
Kalman smoother,
\[
q_j^{(r)}(Z^{(r)})
=
\mc N\!\left(
Z^{(r)};\mu_{0:K}^{(r)},\bs\Sigma_{0:K}^{(r)}
\right),
\]
obtained by locally linearizing the decoder along the filtering and smoothing
trajectory. This choice is natural for K-PWM because the subsequent CM-step depends on the
latent trajectory only through smoothed first moments, marginal covariances, and
lag-one covariances,
\[
\mu_k^{(r)},\qquad
\bs\Sigma_{k,k}^{(r)},\qquad
\bs\Sigma_{k,k+1}^{(r)}.
\]
EKS provides exactly these sufficient moments, which are then used to evaluate
the expected complete-data log likelihood and update the Koopman-bilinear
transition parameters and covariance matrices. Thus, EKS gives a tractable
Gaussian E-step that is consistent with the probabilistic state-space structure
of K-PWM while accommodating the nonlinear observation decoder. In the special case where \(h_{\bs\zeta}\) is affine, the model becomes
conditionally linear-Gaussian and this smoothing step reduces to the standard
Kalman smoother E-step.

\subsection{Why Koopman-bilinear form is advantageous ?}
We refer the reader to Bruder et al. \cite{bruder2021advantages} for a detailed discussion on the advantages of Koopman
bilinear representations for controlled nonlinear systems. The Koopman
framework makes the dynamics linear with respect to the observables, this
linearity does not automatically imply linearity with respect to the control
input. Enforcing linear input dependence typically requires restricting the
Koopman operator to a subspace of observables containing only linear functions of
the input. Finite-dimensional approximations of this restricted operator are
then commonly identified from data using EDMD-based linear system identification
methods. However, such approximations rely on the implicit assumption that
increasing the lifted dimension can recover the true controlled Koopman operator
arbitrarily well. This assumption is not valid in general: a valid Koopman
representation may fail to exist when the observable space excludes nonlinear
functions of the input, even if the state-dependent observable space is
infinite-dimensional. Conversely, every controlled system can be represented by a
Koopman model with nonlinear input dependence, but such representations are often
less useful for control because they sacrifice much of the structure needed for
efficient prediction and controller synthesis. Bilinear Koopman models strike a
balance between these two extremes by allowing the lifted dynamics to remain
linear in the observables while depending multiplicatively on the control input.
This structure is particularly natural for control-affine systems, since the
continuous-time Koopman generator preserves the control-affine dependence in
lifted coordinates, leading to bilinear surrogate dynamics. Consequently,
bilinear Koopman realizations are more expressive than linear-input Koopman
models, more structurally compatible with general controlled nonlinear systems,
and more computationally tractable than fully nonlinear-input realizations. This
makes them especially attractive for model-based control, where one seeks
predictive flexibility without losing the algorithmic benefits of structured
lifted dynamics.

\section{Supplementary Information for \Cref{sec:MPPI}}
\label{sub:mppi_supp}

We use model predictive path integral (MPPI) control as a sampling-based
receding-horizon controller to evaluate whether the learned K-PWM model supports
downstream control under partial observability. At each environment time step
\(k\), the controller maintains a latent belief \((\mu_k,\bs\Sigma_k)\), obtained
from the EKF using the most recent partial observation. Starting from this
belief, MPPI samples \(S\) candidate action sequences over an \(L\)-step planning
horizon,
\[
a_{k:k+L-1}^{(s)}
=
\mc P_{\mc A}
\left(
a_{k:k+L-1}^{\mrm{base}}
+
\epsilon_{k:k+L-1}^{(s)}
\right),
\qquad
s=1,\ldots,S,
\]
where \(a_{k:k+L-1}^{\mrm{base}}\) is a base sequence constructed from the
current nominal sequence and, when available, reference or gait-prior actions.
The term \(\epsilon_{k:k+L-1}^{(s)}\) denotes temporally correlated Gaussian
exploration noise, and \(\mc P_{\mc A}\) denotes projection onto the admissible
action set \(\mc A\), implemented as componentwise clipping to the environment
action bounds in our experiments.

For each sampled action sequence, the controller samples an initial latent state
from the current belief,
\[
\bar z_k^{(s)} \sim \mc N(\mu_k,\bs\Sigma_k),
\]
and rolls out the learned Koopman-bilinear latent dynamics,
\[
\bar z_{k+l+1}^{(s)}
=
\mf A_{k+l}
\bar z_{k+l}^{(s)}
+
\bs b_{k+l},
\qquad
l=0,\ldots,L-1.
\]
The predicted observations are obtained through the learned decoder,
\[
\hat y_{k+l+1}^{(s)}
=
h_{\bs\zeta}
\!\left(
\bar z_{k+l+1}^{(s)}
\right).
\]

\begin{algorithm}[htpb!]
\caption{MPPI Control with Learned \kpwm{}}
\label{alg:mppi_kpwm}

\KwIn{Learned K-PWM parameters $\Theta=\{\widetilde{\Theta},\bs\zeta\}$,
current belief $(\mu_k,\bs\Sigma_k)$, nominal action sequence
$\bar a_{k:k+L-1}$, reference sequence $y_{k+1:k+L}^{\mrm{ref}}$,
planning horizon $L$, number of rollouts $S$, temperature $\lambda$,
discount factor $\gamma$}
\KwOut{Applied action $a_k$ and shifted nominal action sequence
$\bar a_{k+1:k+L}$}

Construct a base action sequence $a_{k:k+L-1}^{\mrm{base}}$ from the current
nominal sequence and, when available, reference or gait-prior actions\;

Sample temporally correlated exploration noise
$\epsilon_{k:k+L-1}^{(s)}$ for $s=1,\ldots,S$\;

Form candidate action sequences
$
a_{k:k+L-1}^{(s)}
=
\mc P_{\mc A}
\left(
a_{k:k+L-1}^{\mrm{base}}
+
\epsilon_{k:k+L-1}^{(s)}
\right),
\quad
s=1,\ldots,S
$\;

Sample initial latent particles
$
\bar z_k^{(s)} \sim \mc N(\mu_k,\bs\Sigma_k),
\quad
s=1,\ldots,S
$\;

Initialize rollout costs $J^{(s)}=0$ for $s=1,\ldots,S$\;

Initialize rollout covariance
$
\bs\Sigma_{k}^{\mrm{roll}}=\bs\Sigma_k
$\;

\For{$l=0,\ldots,L-1$}{

    \For{$s=1,\ldots,S$}{

        Propagate the latent state using the learned Koopman-bilinear model.

        Decode the predicted observation:
        $
        \hat y_{k+l+1}^{(s)}
        =
        h_{\bs\zeta}
        \!\left(
        \bar z_{k+l+1}^{(s)}
        \right)
        $\;

        Accumulate the discounted rollout cost:
        $
        J^{(s)}
        \leftarrow
        J^{(s)}
        +
        \gamma^l
        \ell_{\mrm{MPPI}}
        \!\left(
        \hat y_{k+l+1}^{(s)},
        y_{k+l+1}^{\mrm{ref}},
        a_{k+l}^{(s)},
        a_{k+l-1}^{(s)},
        \bs\Sigma_{k+l}^{\mrm{roll}}
        \right)
        $\;
    }

    Propagate the nominal rollout covariance:
    $
    \bs\Sigma_{k+l+1}^{\mrm{roll}}
    =
    \mf A_{k+l}^{\mrm{base}}
    \bs\Sigma_{k+l}^{\mrm{roll}}
    \left(\mf A_{k+l}^{\mrm{base}}\right)^\top
    +
    \bs\Sigma_w
    $\;
}

Normalize rollout costs:
$
\widetilde J^{(s)}
=
\frac{
J^{(s)}-\min_i J^{(i)}
}{
\mathrm{std}_i(J^{(i)})+\varepsilon
},
\quad
s=1,\ldots,S
$\;

Compute MPPI weights:
$
w^{(s)}
=
\frac{
\exp\!\left(-\widetilde J^{(s)}/\lambda\right)
}{
\sum_{i=1}^{S}
\exp\!\left(-\widetilde J^{(i)}/\lambda\right)
},
\quad
s=1,\ldots,S
$\;

Update the nominal action sequence:
$
\bar a_{k:k+L-1}^{+}
=
\sum_{s=1}^{S}
w^{(s)}
a_{k:k+L-1}^{(s)}
$\;

Apply the first action
$
a_k=\bar a_k^{+}
$\;

Shift the optimized sequence forward:
$
\bar a_{k+1:k+L-1}
\leftarrow
\bar a_{k+1:k+L-1}^{+}
$\;

Append a terminal nominal action to obtain $\bar a_{k+1:k+L}$\;

\Return{$a_k$, $\bar a_{k+1:k+L}$}\;

\end{algorithm}

Each rollout is assigned the cumulative discounted cost
\[
J^{(s)}
=
\sum_{l=0}^{L-1}
\gamma^l
\ell_{\mrm{MPPI}}
\!\left(
\hat y_{k+l+1}^{(s)},
y_{k+l+1}^{\mrm{ref}},
a_{k+l}^{(s)},
a_{k+l-1}^{(s)},
\bs\Sigma_{k+l}^{\mrm{roll}}
\right),
\]
where \(\gamma \in (0,1]\) is the discount factor. The MPPI stage cost
\(\ell_{\mrm{MPPI}}\) contains task-dependent observation-reference tracking
terms, control effort, action smoothness, and an uncertainty penalty. In
particular, the uncertainty term is
\[
\ell_{\mrm{unc}}
=
\lambda_{\mrm{unc}}\,
\mrm{tr}\!\left(\bs\Sigma_{k+l}^{\mrm{roll}}\right),
\]
where the rollout covariance is propagated using the nominal Koopman transition,
\[
\bs\Sigma_{k+l+1}^{\mrm{roll}}
=
\mf A_{k+l}^{\mrm{base}}
\bs\Sigma_{k+l}^{\mrm{roll}}
\left(\mf A_{k+l}^{\mrm{base}}\right)^\top
+
\bs\Sigma_w .
\]
Since \(\bs\Sigma^{\mrm{roll}}\) is propagated along the base sequence, it is the same for
all candidates: the uncertainty term measures the latent uncertainty of the nominal plan
over the horizon and adds a common offset to every \(J^{(s)}\), which the cost normalization
below removes. The candidate weights are therefore determined by the remaining terms of
\(\ell_{\mrm{MPPI}}\) (tracking, control effort and, where used, action smoothness), while the
uncertainty term is part of the reported planning cost.

After evaluating all sampled rollouts, the costs are normalized as
\[
\widetilde J^{(s)}
=
\frac{
J^{(s)}-\min_i J^{(i)}
}{
\mathrm{std}_i(J^{(i)})+\varepsilon
},
\qquad
s=1,\ldots,S.
\]
The normalized costs define exponential MPPI weights,
\[
w^{(s)}
=
\frac{
\exp\!\left(-\widetilde J^{(s)}/\lambda\right)
}{
\sum_{i=1}^{S}
\exp\!\left(-\widetilde J^{(i)}/\lambda\right)
},
\qquad
s=1,\ldots,S,
\]
where \(\lambda>0\) is the MPPI temperature. The nominal action sequence is then
updated by the weighted average
\[
\bar a_{k:k+L-1}^{+}
=
\sum_{s=1}^{S}
w^{(s)}
a_{k:k+L-1}^{(s)}.
\]
Only the first action \(a_k=\bar a_k^{+}\) is applied to the true MuJoCo
environment. The optimized sequence is shifted forward by one step, appended
with a terminal nominal action, and reused as the nominal sequence at the next
control step.

\newpage

\section{Supplementary Material for \Cref{sec:exp}}\label{sub:exp setting}

\paragraph{Computing Resources}\label{sub:comp_res}
All experiments were run on a laptop with an Intel Core i7-11850H CPU, 32 GB RAM, and an NVIDIA GeForce RTX 3080 Laptop GPU with 16 GB VRAM. 

\begin{figure}[htpb!]
    \centering
    \begin{subfigure}{0.95\linewidth}
        \centering
        \includegraphics[width=\linewidth]{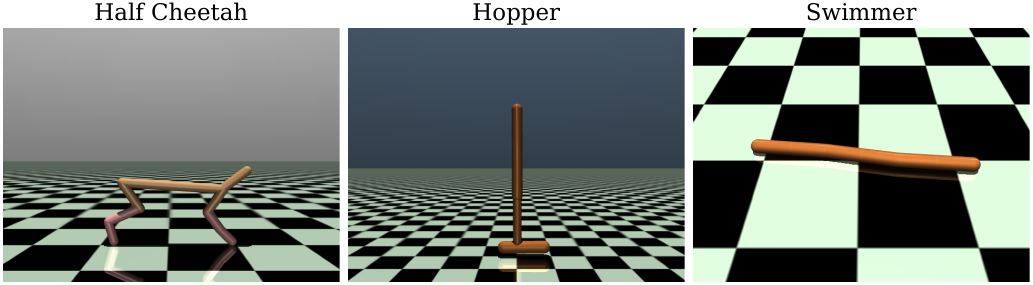}
        \caption{Gymnasium MuJoCo.}
        \label{fig:first}
    \end{subfigure}

    \par\smallskip

    \begin{subfigure}{0.95\linewidth}
        \centering
        \includegraphics[width=\linewidth]{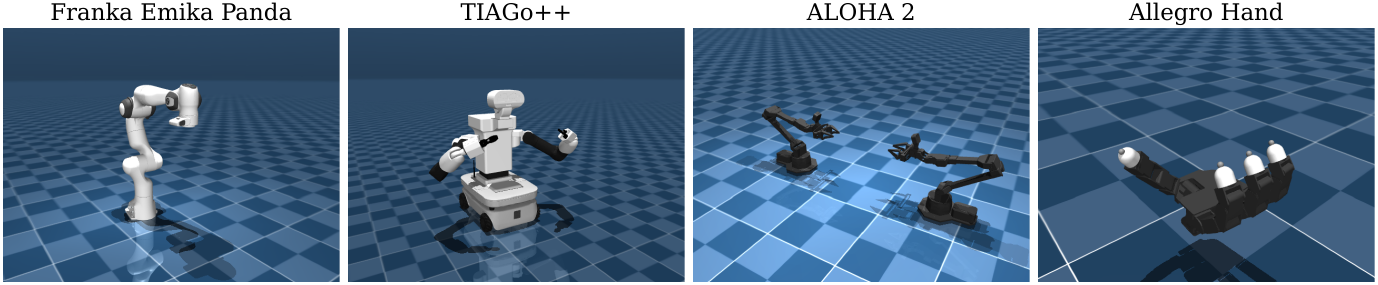}
        \caption{MuJoCo Menagerie.}
        \label{fig:second}
    \end{subfigure}

    \caption{Two sets of systems in MuJoCo experiments considered.}
    \label{fig:exp_systems}
\end{figure}

\paragraph{Systems considered in experiment.} As already mentioned in the main text, we consider two sets of systems based-on  Gymnasium MuJoCo \footnote{\url{https://gymnasium.farama.org/environments/mujoco/}} and DeepMind's MuJoCo Menagerie\footnote{\url{https://mujoco.readthedocs.io/en/stable/models.html}} as shown in \Cref{fig:exp_systems}.

\subsection{Training Data Collection}

For every task we simulate a pool of $150$ trajectories of actions and observations. The candidate trajectory $r$ of the system $s$ uses the
seed $s + r$, which fixes its random input sequence, its initial state, and its
measurement noise. Before simulation, the input sequence is tested for PE condition \eqref{eq:pe_test} with $\lambda_{\mrm{PE}} = 0.02$. A
candidate action sequence that fails the PE test, or whose rollout is invalid (e.g., a fall for Hopper, any contact
for the free-space robots), is discarded and the next seed is used until $150$ trajectories are accepted. Every observation channel is corrupted at every time step by
independent Gaussian noise $\nu_k \sim \mathcal N(0, \sigma^2_o)$, and all models are trained on
the noisy observations. For each system, we draw disjoint training, validation, and test sets of 15, 10, and 10 trajectories, respectively, from the pool of 150 candidate trajectories. These drawn trajectories are held fixed for all the methods. In the next, $\Delta t$ represent sampling time.

\begin{figure}[t]
    \centering
    \includegraphics[width=0.82\linewidth]{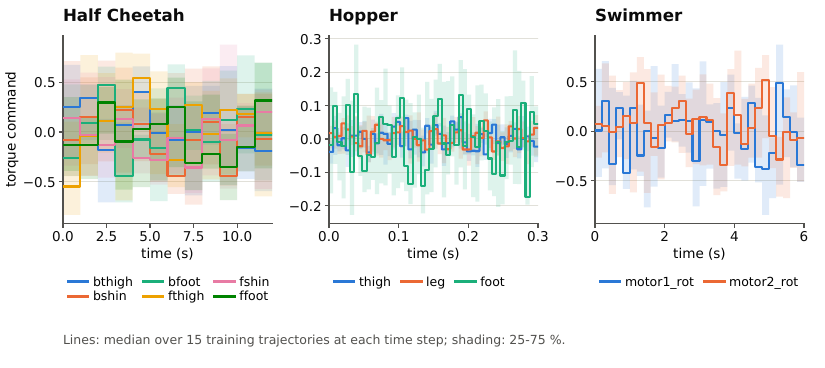}
    \caption{The actions with time of the training trajectories for the Gymnasium MuJoCo systems.}
    \label{fig:gym_training_actions}
\end{figure}

\paragraph{Gymnasium tasks.}
Half Cheetah, Hopper and Swimmer (Gymnasium v5), started from Gymnasium's seeded reset.
Actions are piecewise-constant torques, each drawn uniformly from $[-\bar a, \bar a]$ and
held for a fixed interval:
Half Cheetah ($\bar a = 1$, held $1$\,s, for $12$\,s at $\Delta t = 0.1$\,s),
Swimmer ($\bar a = 1$, held $0.2$\,s, for $6$\,s at $\Delta t = 0.1$\,s) and
Hopper ($\bar a = 0.1, 0.1, 0.3$ for thigh, leg and foot; held $6$\,ms; for $0.3$\,s at
$\Delta t = 2$\,ms; as larger torques make the Hopper fall).
We use $\sigma_o = 0.01$  for Half Cheetah and Swimmer, and
$\sigma_o = 0.001$ for Hopper. The partial observation setting for the Gymnasium systems considered is provided in \Cref{tab:control_partial_observation_setting}. The actions with time for the training trajectories are shown in \Cref{fig:gym_training_actions}. 

\begin{table}[htpb!]
\centering
\caption{Partial-observation setting for Gymnasium MuJoCo systems.}
\label{tab:control_partial_observation_setting}
\small
\setlength{\tabcolsep}{4pt}
\begin{tabular}{lcccccc}
\toprule
Environment
& $n_y^{\rm full}$
& $n_y$
& Masked
& Observed indices
& Masked indices
& $(\Delta t, T)$ \\
\midrule
HalfCheetah
& 17 & 9 & 8
& $[0{:}8]$
& $[9{:}16]$
& $(0.1,\,12.0)$ \\

Hopper
& 11 & 7 & 4
& $[0{:}6]$
& $[7{:}10]$
& $(0.002,\,0.3)$ \\

Swimmer
& 10 & 7 & 3
& $[0{:}4]\cup[8{:}9]$
& $[5{:}7]$
& $(0.1,\,6.0)$ \\
\bottomrule
\end{tabular}
\end{table}

\paragraph{MuJoCo Menagerie tasks.}

\begin{figure}[htpb!]
    \centering
    \includegraphics[width=0.70\linewidth]{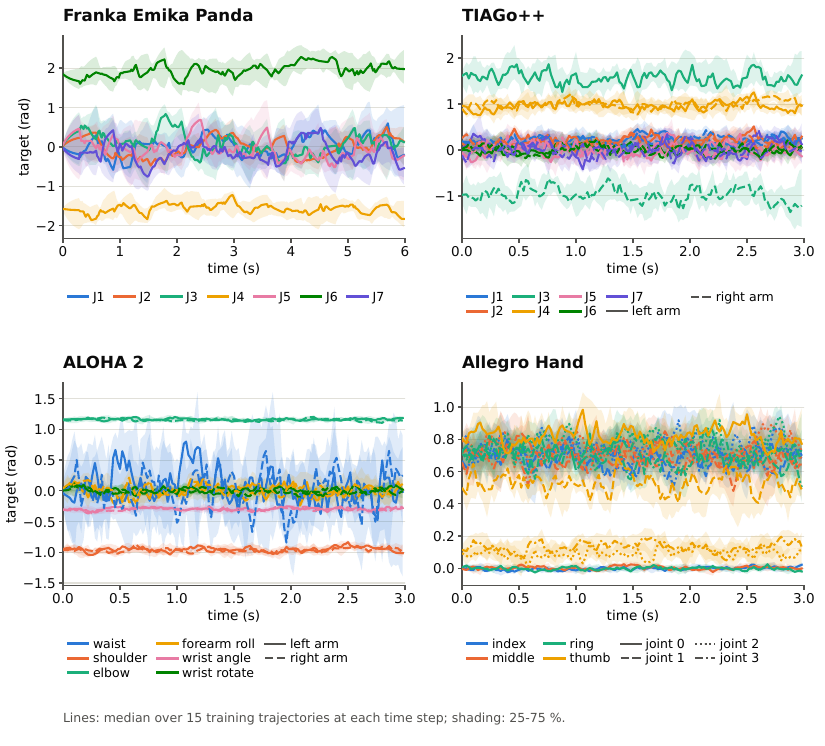}
    \caption{The actions with time of the training trajectories for the MuJoCo Menagerie systems.}
    \label{fig:men_training_actions}
\end{figure}

We simulate each robot with a time step of $\Delta t = 0.02$,s. Trajectories last $6$ s for Panda and $3$ s for the other robots. We sample initial configurations uniformly from the middle $40\%$ of the joint range for Panda and TIAGo++ and from the middle $60\%$ of the actuator range for ALOHA~2 and Allegro.
The actions are joint-position commands. We sample a new target $a^{\star}$ uniformly every $0.3$ s for Panda and every $0.06$ s for the other robots. For Panda and TIAGo++, we restrict the targets to $60\%$ of the actuator range. For ALOHA~2 and Allegro, we use the full range. At each time step, we move the command toward the current target using the first-order filter $a_{k+1} = a_k + \alpha(a^{\star} - a_k)$, with $\alpha = 0.1$ for Panda and $\alpha = 0.2$ for the other robots. We observe joint angles, joint velocities, and the Cartesian positions of the end effectors or fingertips. The resulting observation dimensions are $n_y = 17$, $34$, $30$, and $44$ for Panda, TIAGo++, ALOHA~2, and Allegro, respectively. We discard trajectories in which any contact occurs and set the observation noise level to $\sigma_o = 0.005$. The actions with time for the training trajectories are shown in \Cref{fig:men_training_actions}. 

\subsection{Experiment Setting}\label{sub:exp_set}

\paragraph{Exclusion of Otto et al. \cite{otto2024learning} baseline.}
We attempted to evaluate Otto et al.~\cite{otto2024learning}, the closest partially observed Koopman baseline, on the MuJoCo Menagerie tasks. We used the same training data and latent dimensions selected for \kpwm{}, while retaining the random generator initialization prescribed in~\cite{otto2024learning}. In these experiments, the EM iterations failed to converge, producing non-finite generator estimates or degenerate covariance matrices that prevented subsequent inference and control evaluation. A possible contributing factor is the interaction between random initialization and the affine observation model: the latent dynamics must accommodate observation structure that \kpwm{} can represent through its nonlinear decoder, potentially increasing sensitivity to the initial latent representation and the conditioning of the EM updates. However, the observed failures do not establish a unique cause or imply that the method cannot converge under alternative configurations. We therefore omit this baseline from the reported Menagerie comparison and interpret these results as a numerical limitation of the tested configuration.

\paragraph{Initialization of GECM parameters.}
For the Koopman-consistent initialization of the \kpwm{} parameters we use the same
procedure and hyperparameters for both task families: $\hat R = 10$ randomly selected
training trajectories, a delay window of $\tau = 10$ samples, Legendre features of maximum
degree $d_p = 3$ and generator ridge $\lambda_g = 10^{-6}$; the observation-noise covariance
is initialized as $\bs\Sigma_v = 0.01\,\mf I$. The latent dimension $N = d_p\,r_e$ follows from
the Hankel spectrum through the energy threshold $\rho$, which is the only task-specific
initialization parameter: for the Gymnasium tasks $\rho = 0.8$ (HalfCheetah), $0.95$
(Hopper) and $0.7$ (Swimmer), giving $N = 9$, $6$ and $9$; for the Menagerie robots
$\rho = 0.6$ (Franka Emika Panda, Allegro Hand), $0.99$ (TIAGo++) and $0.7$ (ALOHA~2),
giving $N = 15$, $36$, $234$ and $18$. The MLP decoder has input dimension $N$, one hidden
layer of width $2N$ with ReLU activations, and output dimension $n_y$; its parameters
$\bs\zeta$ use the default PyTorch initialization for linear layers. 

\paragraph{During GECM training.}
During GECM the
closed-form updates are regularized by a generator ridge $\gamma_G$ and eigenvalue floors
$\gamma_w, \gamma_v, \gamma_0$ on $\bs\Sigma_w, \bs\Sigma_v, \bs\Sigma_0$: $\gamma_G = 10^{-2}$
for HalfCheetah, $10^{-4}$ for TIAGo++, $10^{-6}$ for Allegro Hand and $10^{-3}$ otherwise, with
$\gamma_w = \gamma_v = \gamma_0 = 10^{-6}$ for Panda and Allegro Hand and $10^{-5}$
otherwise. Each decoder update (CM-step 2) runs $1000$ Adam epochs at learning rate
$10^{-2}$.
 In CM-step~2 of GECM iteration \(j\), starting from \(\bs\zeta_{j-1}\), the decoder is fitted to the
pairs \((\mu_k^{(r)}, y_k^{(r)})\) of smoothed latent means and observations,
\(k=0,\ldots,K\), \(r=1,\ldots,R\), by minimizing the mean-squared reconstruction error with
full-batch \texttt{Adam} for \(1000\) epochs at learning rate \(10^{-2}\) (decayed by a factor of
\(0.9999\) every \(100\) epochs), on a random \(80\%\) of the pairs; the remaining \(20\%\) are
only monitored. 

\paragraph{GECM training convergence and stopping criteria.}
After each GECM iteration \(j\), we evaluate the regularized training log-likelihood per
observed scalar,
\[
\ell_j
=
\frac{\mathcal L(\Theta_j)-\mcr R(\widetilde\Theta_j)}{R\,(K+1)\,n_y},
\]
where \(\mathcal L\) is the observed-data log-likelihood \eqref{eq:LL} of the \(R\) training
trajectories and \(\mcr R\) is the regularizer of the regularized ELBO. Iteration \(j\) becomes
the incumbent, and its parameters \(\Theta_j\) are stored, only if
\(\ell_j > \ell^\star + \epsilon\), where \(\ell^\star\) is the value of the current incumbent and
\(\epsilon = 10^{-3}\) nats. Training is considered converged, and stops, once \(10\)
consecutive iterations fail to improve on the incumbent by more than \(\epsilon\) (Patience $P=10$); this rule is
applied only after a minimum of \(20\) iterations, and training is capped at
\(j_{\mrm{max}} = 200\). The incumbent parameters are returned. 

\paragraph{MPPI Implementation.}
For the Menagerie robots, the observation stacks the joint angles $q\in\mathbb R^{n_q}$, the
joint velocities $\dot q\in\mathbb R^{n_q}$ and the Cartesian positions $p_e\in\mathbb R^3$ of the
$n_e$ tracked end effectors (fingertips for the Allegro Hand),
$y = [\,q^\top\ \ \dot q^\top\ \ p_1^\top \cdots p_{n_e}^\top\,]^\top$. For candidate $s$ at horizon
step $l$, the decoded prediction
$\hat y_{k+l+1}^{(s)} = [\,\hat q^\top\ \ \hat{\dot q}^\top\ \ \hat p_1^\top \cdots \hat p_{n_e}^\top\,]^\top$
is compared with the reference $y_{k+l+1}^{\mrm{ref}}$ through the stage cost
\[
\begin{aligned}
\ell_{\mrm{MPPI}}
={}&
\lambda_{\mrm{ee}} {\textstyle\sum}_{e=1}^{n_e}
\big\| \hat p_e - p_e^{\mrm{ref}} \big\|_2^2
+
\tfrac{\lambda_q}{n_q} \big\| \hat q - q^{\mrm{ref}} \big\|_2^2
+
\tfrac{\lambda_{\dot q}}{n_q} \big\| \hat{\dot q} - \dot q^{\mrm{ref}} \big\|_2^2 \\
&+
\lambda_a \big\| a_{k+l}^{(s)} \big\|_2^2
+
\lambda_{\mrm{unc}}\, \mrm{tr}\!\big(\bs\Sigma_{k+l+1}^{\mrm{roll}}\big),
\end{aligned}
\]
with $\lambda_{\mrm{ee}} = 800$, $\lambda_q = 20$, $\lambda_{\dot q} = 2$, $\lambda_a = 0.01$ and
$\lambda_{\mrm{unc}} = 0.001$. The end-effector term carries the tracking task; the joint terms keep
the robot near the reference configuration, which the end-effector positions alone do not
determine for redundant arms; the control term penalizes large commands; and the uncertainty
term penalizes the latent covariance propagated along the rollout,
$\bs\Sigma_{k}^{\mrm{roll}} = \bs\Sigma_k$ and
$\bs\Sigma_{k+l+1}^{\mrm{roll}} = \mf A_{k+l}^{\mrm{base}}\bs\Sigma_{k+l}^{\mrm{roll}}(\mf A_{k+l}^{\mrm{base}})^\top + \bs\Sigma_w$,
with $\mf A_{k+l}^{\mrm{base}}$ the Koopman-bilinear transition under the base action $a_{k+l}^{\mrm{base}}$.
The rollout cost is
$J^{(s)} = \sum_{l=0}^{L-1}\gamma^l \ell_{\mrm{MPPI}}$ with $\gamma = 0.98$, and a non-finite cost is
replaced by a large constant before normalization. The base sequence is
$a^{\mrm{base}}_{k:k+L-1} = 0.85\,a^{\mrm{ref}}_{k:k+L-1} + 0.15\,\bar a_{k:k+L-1}$, clipped to the
actuator range, where $\bar a$ is the shifted nominal sequence; the exploration noise follows
$\epsilon_l = 0.85\,\epsilon_{l-1} + 0.15\,\eta_l$ with $\eta_l\sim\mc N(0, 0.15^2\mf I)$; and the action
appended after the shift is $a^{\mrm{ref}}_{k+L}$. The cost weights, base
sequence and exploration noise are identical for every world model. MPPI uses $S = 2000$ candidates, a horizon of $L = 20$ steps ($0.4$\,s at $\Delta t = 0.02$\,s) and temperature $\lambda = 0.7$. The actions are
joint-position commands ($n_q = 7$, $14$, $12$ and $16$ for Panda, TIAGo++, ALOHA~2 and Allegro
Hand), and the robots track $n_e = 1$, $2$, $2$ and $4$ end effectors, respectively. Each
closed-loop run tracks a $5.58$\,s circular reference ($279$ control steps), starting from the
home configuration with every joint offset by $\mc U[-0.01, 0.01]$\,rad; the controller observes
all channels corrupted by zero-mean Gaussian noise with standard deviation $0.005$, while
tracking errors are measured on the noise-free state. Results are reported over five runs
(seeds $0$--$4$), each seed fixing the start offset, the measurement noise and the MPPI
sampling; all world models see the same seeds.

\paragraph{RMSE metric.}
For each prediction horizon, RMSE is computed per test trajectory and then
summarized across trajectories. Specifically, for prediction horizon $s$, let
$\widehat y_{i,s}\in\mathbb{R}^{n_y}$ and $y_{i,s}\in\mathbb{R}^{n_y}$ denote the
predicted and true observations for test trajectory $i$. The per-trajectory RMSE
is
\[
\mathrm{RMSE}_i(s)
=
\sqrt{
\frac{1}{n_y}
\sum_{d=1}^{n_y}
\left(
\widehat y_{i,s,d}-y_{i,s,d}
\right)^2
}.
\]
The reported RMSE is the mean over the $N_{\mathrm{test}}$ test trajectories,
\[
\overline{\mathrm{RMSE}}(s)
=
\frac{1}{N_{\mathrm{test}}}
\sum_{i=1}^{N_{\mathrm{test}}}
\mathrm{RMSE}_i(s),
\]
and the reported standard deviation is the sample standard deviation across
test trajectories,
\[
\mathrm{Std}(s)
=
\sqrt{
\frac{1}{N_{\mathrm{test}}-1}
\sum_{i=1}^{N_{\mathrm{test}}}
\left(
\mathrm{RMSE}_i(s)
-
\overline{\mathrm{RMSE}}(s)
\right)^2
}.
\]
Thus, the uncertainty reported with RMSE reflects variability across test
trajectories rather than the standard error of the mean.

\paragraph{End-effector tracking RMSE.}
For each robot, we evaluate end-effector position tracking
against the reference trajectory using the noise-free simulated
positions. Let $E_r$ denote the end-effector tracking RMSE,
expressed in millimeters, for independent run $r$.
Table~2 reports the mean and sample standard deviation over
$N_{\mathrm{run}}=5$ runs:
\[
\bar{E}
=
\frac{1}{N_{\mathrm{run}}}
\sum_{r=1}^{N_{\mathrm{run}}} E_r,
\qquad
s_E
=
\sqrt{
\frac{1}{N_{\mathrm{run}}-1}
\sum_{r=1}^{N_{\mathrm{run}}}
(E_r-\bar{E})^2
}.
\]
Thus, the reported values are $\bar{E}\pm s_E$, with the
standard deviation measuring variability across independent
runs. Joint-angle and joint-velocity errors are excluded
from this metric.

\newpage

\end{document}